\documentclass{article}

\usepackage{microtype}
\usepackage{graphicx}
\usepackage{subfigure}
\usepackage{booktabs} 
\usepackage{xurl} 

\usepackage{hyperref}

\usepackage[accepted]{arxiv}

\usepackage{amsmath}
\usepackage{amssymb}
\usepackage{mathtools}
\usepackage{amsthm}
\usepackage{booktabs}

\usepackage[capitalize,noabbrev]{cleveref}

\usepackage{tikz} 
\usepackage{pgfplots}
\pgfplotsset{compat=1.17}

\usepgfplotslibrary{polar}
\usetikzlibrary{calc} 
\usetikzlibrary{positioning}

\usepackage{radarchart}

\usepackage{verbatim}

\theoremstyle{plain}

\theoremstyle{definition}

\theoremstyle{remark}

\usepackage[textsize=tiny]{todonotes}

\usepackage{multirow}
\usepackage{array}

\usepackage{booktabs}    
\usepackage{adjustbox}   
\usepackage{graphicx}    
\usepackage{multirow}    
\usepackage{colortbl}    
\usepackage{xcolor}      
\usepackage{diagbox}     

\usepackage{hyperref}    

\definecolor{gray}{gray}{0.9} 

\arxivtitlerunning{AdaPRL: Adaptive Pairwise Regression Learning with Uncertainty Estimation for Universal Regression Tasks}

\begin{document}

\twocolumn[
\arxivtitle{AdaPRL: Adaptive Pairwise Regression Learning with Uncertainty Estimation for Universal Regression Tasks}

\arxivsetsymbol{equal}{*}

\begin{arxivauthorlist}
\arxivauthor{Fuhang Liang}{comp}
\arxivauthor{Rucong Xu}{uni,comp}
\arxivauthor{Deng Lin}{comp}

\end{arxivauthorlist}

\arxivaffiliation{comp}{Tencent Music Entertainment, Shenzhen 518057, China}
\arxivaffiliation{uni}{Shenzhen Institute for Advanced Study, University of Electronic Science and Technology of China, Shenzhen 518110, China}

\arxivcorrespondingauthor{Fuhang Liang}{frankyliang@tencent.com}

\arxivkeywords{Machine Learning, arxiv}

\vskip 0.3in
]



\printAffiliationsAndNotice{}  

\begin{abstract}

Current deep regression models usually learn in a point-wise way that treats each sample as an independent input, neglecting the relative ordering among different data. Consequently, the regression model could neglect the data's interrelationships, potentially resulting in suboptimal performance. Moreover, the existence of aleatoric uncertainty in the training data may drive the model to capture non-generalizable patterns, contributing to increased overfitting. To address these issues, we propose a novel adaptive pairwise learning framework for regression tasks (AdaPRL) which leverages the relative differences between data points and integrates with deep probabilistic models to quantify the uncertainty associated with the predictions. Additionally, we adapt AdaPRL for applications in multi-task learning and multivariate time series forecasting. Extensive experiments with several real-world regression datasets including recommendation systems, age prediction, time series forecasting, natural language understanding, finance, and industry datasets show that AdaPRL is compatible with different backbone networks in various tasks and achieves state-of-the-art performance on the vast majority of tasks without extra inference cost, highlighting its notable potential including enhancing prediction accuracy and ranking ability, increasing generalization capability, improving robustness to noisy data, improving resilience to reduced data, and enhancing interpretability. Experiments also show that AdaPRL can be seamlessly incorporated into recently proposed regression frameworks to gain performance improvement.

\end{abstract}

\section{Introduction}
\label{introduction}

In recent years, regression tasks have become increasingly important in various domains, including computer vision, time series forecasting, recommendation systems, finance, and so on. Traditional regression methods which integrate with point-wise loss functions such as mean squared error (MSE), mean absolute error (MAE) and Huber have been widely adopted due to their simplicity and effectiveness. However, these traditional pointwise methods often struggle with capturing the nuanced relationships between data points, potentially resulting in less effective performance. Pairwise learning, on the other hand, offers a promising alternative by focusing on relative differences between pairs of data points rather than on absolute values. 

\begin{figure}[ht]
\vskip 0.1in
\begin{center}
\begin{minipage}{1.0\columnwidth}
\centerline{\includegraphics[width=1\columnwidth]{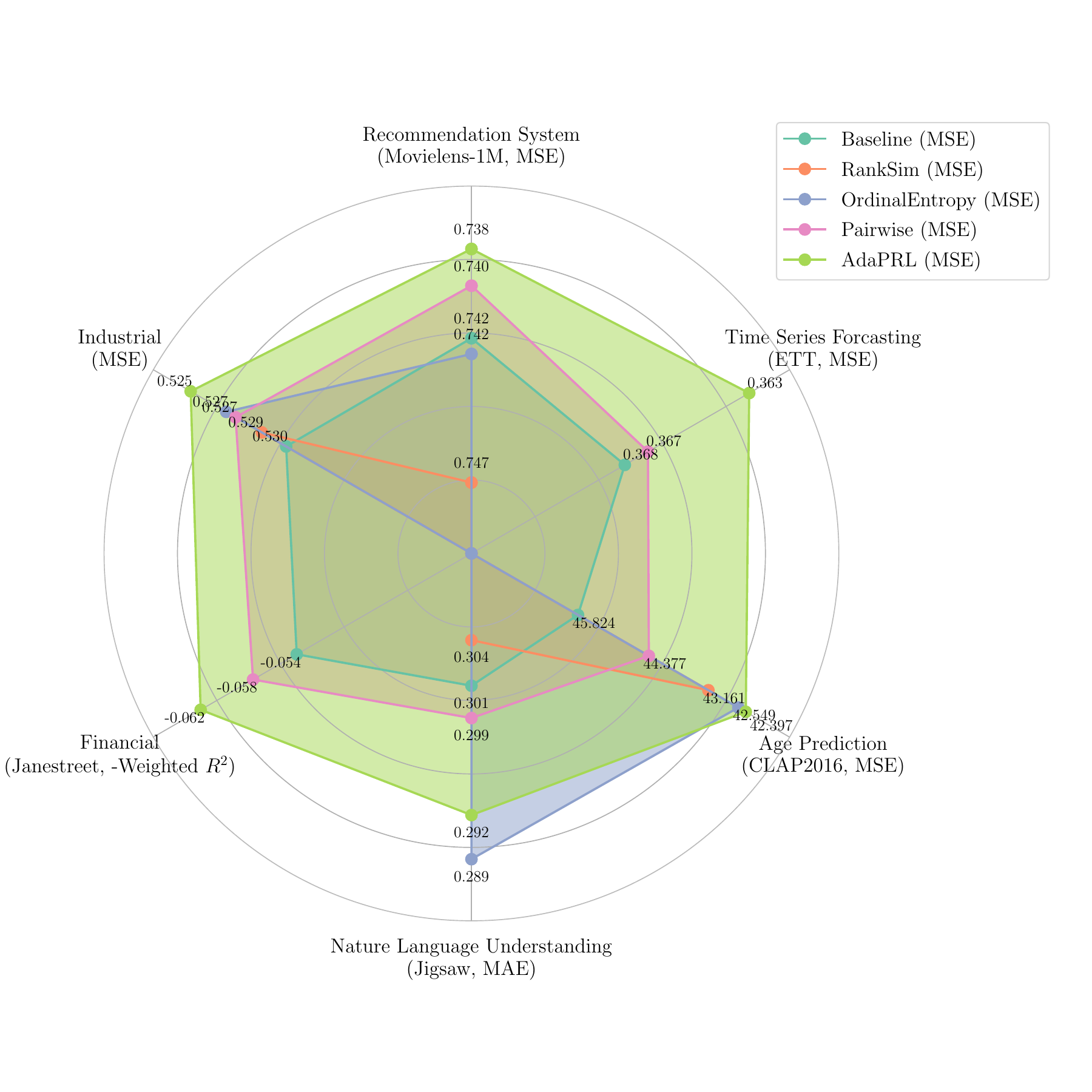}}
\caption{Overall performance of AdaPRL and comparison with different regression loss on certain metrics with different dataset across multiple domains.}
\label{radar}
\end{minipage}
\end{center}
\vskip -0.1in
\end{figure}

To overcome the limitations of point-wise learning, Learning-to-Rank (LTR) approaches have been introduced, where pairwise methods aim to ensure that the predicted score of a positive example surpasses that of a negative sample for each positive-negative pair, and listwise methods promote higher rankings for positive samples across the entire list of items. Representative pairwise learning methods include RankSVM \cite{joachims2002optimizing}, GBRank \cite{zheng2007regression}, RankNet \cite{burges2005learning}, LambdaRank \cite{burges2006learning} and LambdaMART \cite{burges2010ranknet}. Some previous studies \cite{li_click-through_2015,yan_scale_2022,yue_learning_2022,sheng_joint_2023} have investigated that combining binary cross-entropy loss and auxiliary ranking loss improves classification performance. A more recent work \cite{lin_understanding_2024} has also shown that the combination of binary cross-entropy loss and ranking loss can address the gradient vanishing issue for negative samples, especially in the context of sparse positive feedback scenarios, thereby improving the classification performance.

While existing literature has shown that the combination of pairwise learning with binary cross-entropy loss in classification problems enhances classification accuracy, only a few studies \cite{he2024rankability} have studied combining pairwise learning with the regression problem to our knowledge, yet they mainly focus on uplift modeling rather than on regression problems. Meanwhile, some recent studies \cite{dai2023semi,zhou2024ranking,NEURIPS2023_39e9c591,cheng2024appearance} try to employ contrastive learning to enhance regression tasks by improving feature representations through leveraging sample similarities and preserving label distance relationships within the feature space. However, these researches are mainly concentrated in the field of image processing instead of proposing generalized methods for tabular data and other real-world tasks.

Furthermore, noise and uncertainty in the dataset adversely affect the regression algorithm's learning process, leading to a decrease of accuracy and robustness, yet less attention is paid to the issue. Uncertainty estimation is the process of quantifying the uncertainty associated with the predictions made by machine learning models. Traditional pointwise predictions provide a single value as an estimate without conveying the range of possible outcomes or their likelihood, whereas probabilistic predictions predict a distribution of possible values, enabling better decisions under uncertainty. Early work on uncertainty estimation formalizes the task as a Bayesian problem by assigning a prior distribution to the parameters of the statistical model and updating the distribution of the parameters conditional on the data to estimate predictive uncertainty \cite{bernardo2009bayesian}. Some approximated Bayesian methods \cite{mackay1992bayesian,neal2012bayesian,blundell2015weight,graves2011practical} are proposed to alleviate the computation bottleneck. Machine learning models such as generalized additive models \cite{rigby2005generalized}, random forests \cite{meinshausen2006quantile} and boosting algorithms \cite{hofner2014model,duan2020ngboost} can also predict specific quantiles of the predictive distribution accurately using quantile regression \cite{meinshausen2006quantile,bhat2015towards} or distribution regression \cite{schlosser2019distributional,thomas2018gradient}. The ensemble methods \cite{hall2007combining,kapetanios2015generalised,bassetti2018bayesian} have been shown to be effective in uncertainty estimation or density estimation. Monte Carlo dropout \cite{gal2016dropout} is also proposed to estimate predictive uncertainty. The presence of aleatoric uncertainty, as we will demonstrate later, also affects the learning process of regular pairwise learning.

In this paper, we address the challenge of improving regression tasks by integrating pairwise learning and uncertainty learning using deep probabilistic models. Specifically, we formulate the regression problem as a combination of the original regression task together with an auxiliary uncertainty-based learning task; not only does the model generate a more accurate point-wise prediction, but it also predicts the relative differences between pairs of data points with the uncertainty associated with these predictions. This formulation allows the model to capture the underlying relationships and dependencies within the data more effectively and provides more reliable and robust predictions in a wide range of regression tasks.

The motivation for our research stems from the limitations of existing regression methods, especially in handling complex and high-dimensional data, as well as quantifying the uncertainty of the data. We propose a new method called \textbf{Adaptive Pairwise Regression Learning (AdaPRL)}. By integrating uncertainty learning, our method can estimate the aleatoric uncertainty level for each sample, allowing our pairwise regression learning framework to focus more on sample pairs with higher certainty instead of unreliable data points. We also propose adaptation of AdaPRL so that it could work not just with single output prediction but also multi-task learning and multivariate time series prediction. Our research has conducted extensive comparative experiments on different regression datasets in a wide range of fields including but not limited to recommendation systems, time-series forecasting, computer vision, and financial and industrial areas, which shows that our proposed AdaPRL framework not only improves the accuracy and robustness of the original model but also enhances the ranking ability of the model. 

\textbf{Contributions} 

The core contributions of our work include:
\begin{enumerate}
    \item We propose a novel pairwise learning framework for regression tasks that leverages the relative differences between data points to improve prediction accuracy with two different loss types which integrates deep probabilistic models to quantify the uncertainty associated with the predictions. This integration allows us to provide more reliable and robust predictions. 
    \item We introduce a new loss function that combines the strengths of pairwise learning with traditional regression loss functions, such as MSE and MAE, and incorporates uncertainty estimation. This hybrid loss function is designed to capture both the absolute and relative aspects of the data while accounting for uncertainty. 
    \item We conduct extensive experiments on a variety of regression datasets, including image datasets (e.g., AgeDB and CLAP2016), time series datasets (e.g., ETT and Electricity), natrue language understanding datasets (e.g., Jigsaw),  recommendation systems datasets (e.g., MovieLens-1M and KuaiRec), financial dataet (e.g., Janestreet) and a large-scale proprietary industrial data from a major music streaming app. Extensive experiment results demonstrate that our proposed method can be combined with various backbone models and outperforms the original regression loss on different datasets. Further experiment in the Appendix shows that our method also has the potential to increase generalization ability, robustness and interpretability of modern regression model. Our proposed AdaPRL can also be well integrated with existing regression methods such as RankSim \cite{gong2022ranksim} and Ordinal Entropy \cite{zhang2023improving}.
\end{enumerate}

\section{Methodology}

\subsection{Problem Setup}
In deep learning regression tasks, the primary objective is to train a neural network model using specific regression loss functions. The neural network typically consists of a feature encoder $ f(\cdot): \mathcal{X} \rightarrow \mathbb{R}^{d_e} $ and a predictor $ g(\cdot): \mathbb{R}^{d_e} \rightarrow \mathbb{R}^{d_t} $. The predictor is designed to estimate the target vector $ \mathbf{y} \in \mathbb{R}^{d_t} $ based on the encoded features $f(\mathbf{x})$ obtained from the input data $\mathbf{x} \in \mathcal{X}$. 

\subsection{Adaptive Pairwise Regression Learning (AdaPRL)}
In this session, we analyze the regression problem from a pairwise learning perspective and propose an innovative pairwise ranking framework for regression called Adaptive Pairwise Regression Learning (AdaPRL), where the regression models not only predict the target of the data points but also learn the relative rankings and certainty of the sample pairs. This helps the regression model to be aware of the prediction relativity, thus improving the accuracy as well as the ranking ability. To conduct a comprehensive analysis of the problem, we address several key issues: the definition of positive and negative pairs, the optimization process, the impact of data noise on pairwise learning, and how our proposed method mitigates this noise effect.

\subsubsection{Definition of the positive and negative examples}

Unlike classification problems, the way to define positive and negative sample pairs has been fully studied in previous work \cite{li_click-through_2015,yan_scale_2022,yue_learning_2022,sheng_joint_2023}. For example, in the binary classification scenario, the positive or negative nature of samples is typically defined directly by their binary labels. We want the model prediction for positive samples to be higher than for negative samples. Therefore, when one sample is of the positive class and the other is of the negative class, we can naturally expect the model's predicted score for the positive class sample to be higher than that for the negative class sample \cite{li_click-through_2015}. In multi-class classification settings, a prevalent approach for defining positive and negative samples in contrastive learning involves designating samples from the same class or identity as positive pairs and samples from different classes or identities as negative pairs \cite{radford2021learning}, which facilitates the model's ability to learn discriminative representations by minimizing the distance between positive pairs while maximizing the distance between negative pairs. Meanwhile, data augmentation and transformation techniques are frequently used to define positive and negative samples in earlier research on contrastive learning \cite{chen2020simple,he2020momentum,grill2020bootstrap,NEURIPS2023_39e9c591}.

However, the definition of positive and negative samples is not as straightforward in the settings for regression tasks. In the context of regression problems, which typically involve predicting unbounded and continuous labels, it is difficult to directly define positive and negative samples solely based on the absolute magnitude of the labels because a sample with a large magnitude might still be relatively small when compared to another sample; therefore, we took an alternative approach to define positive-negative sample pairs based on the relative magnitude of the samples' labels. 

Specifically, for a given pair of samples $(i, j)$, first we need to calculate the difference of labels:

\begin{equation}
\Delta y_{ij} = y_i - y_j.
\end{equation}

We follow \cite{li_click-through_2015,10.1145/2501040.2501978} to incorporate hinge loss with pairwise learning. For convenience, assume a threshold parameter \(\theta\) and define a pair of samples as a positive-negative pair if and only if the difference between the labels is greater than or equal to the threshold \(\theta\) and define a boolean matrix \(M\) with size \(i \times j\) to represent if \(\Delta y_{ij}\) is greater than \(\theta\), that is to say:

\begin{equation}
M_{ij} = \begin{cases} 
1 & \text{if } y_i - y_j > \theta, \label{eq_2}\\
0 & \text{otherwise}.
\end{cases}
\end{equation}

Therefore, the non-zero elements of $M$ represent valid positive-negative pairs. 

\subsubsection{Optimization with auxiliary pairwise ranking loss in regression task}
To optimize the model based on the auxiliary pairwise ranking loss (PRL), we need to define a pairwise ranking loss function. Similar to a recent work on uplift modeling with pairwise learning 
 \cite{he2024rankability}, a straightforward way of combining pairwise learning with regression tasks is to make the difference between two samples' predictions as close as possible to the difference of their labels. To achieve that, firstly calculate the difference of the model prediction of an arbitrary data pair of sample \(i\) and sample \(j\) as:
\begin{equation}
\Delta s_{ij} = \hat{y_i} - \hat{y_j}.
\end{equation}

Secondly, calculate pairwise loss \( L_{prl} \) according to the predefined loss type. Since batch gradient descent is commonly used in practice, we focus on pairwise ranking learning with data batches. Similar to standard regression tasks, where mean squared error (MSE) and mean absolute error (MAE) have been widely applied, the optimization target for the pairwise regression framework can also be formalized by treating all errors either linearly or quadratically. We refer to the former as the \emph{MAE} type and the latter as the \emph{RMSE} type for the purpose of simplicity.

Specifically, if \emph{MAE} type pairwise learning loss is used, then pairwise regression loss becomes:
\begin{equation}
   L_{prl-mae}=\frac{1}{D} \sum_{i=1}^{B}\sum_{j=1}^{B}M_{ij} |\Delta s_{ij} - \Delta y_{ij}|.\label{eq_4}
\end{equation}
If \emph{RMSE} type pairwise learning loss is used, then:
\begin{equation}
  L_{prl-rmse} = \sqrt{\frac{1}{D} \sum_{i=1}^{B}\sum_{j=1}^{B} M_{ij} (\Delta s_{ij}- \Delta y_{ij})^2},\label{eq_5}
\end{equation}
where \(B\) is mini-batch size, and \(D\) is the number of true elements of the hinge matrix M. We will refer to the method as pairwise regression learning for brevity in subsequent sections.

\subsubsection{AdaPRL: Tackling noise in pairwise learning by uncertainty estimation}

Aleatoric uncertainty is inherent and represents a common challenge in real-world applications, which can arise from ambiguities in the data or labels, errors in data annotation and collection, intrinsic randomness of the data, or random errors from sensors or cameras used for data collection \cite{hullermeier2021aleatoric}. While the standard pairwise learning loss proposed above contributes to improving the accuracy and the ranking ability of regression models, which we will demonstrate in the experiments, it is important to acknowledge that this straightforward approach has a major drawback when data noise or uncertainty is taken into account. 

From a distributional perspective of regression tasks, the target $y$ can be considered as a random sample drawn from the dataset $D$, where $y$ follows the conditional probability distribution \( p(y|x;D) \). Consequently, for any pair of samples $(i, j)$ that has a high probability of containing data noise, the ranking and relative magnitude of this pair are also likely to be noisy. Therefore, forcing the difference of their predicted values to be equal to the label difference \( \Delta y_{ij} \) during training, as described above, could negatively impact the precision of both the ranking and regression tasks, exacerbate model overfitting, or even cause mode collapse.

To further explain the phenomenon, consider the following example where random noise cannot be ignored. Assume that both samples $i$ and $j$ contain a certain level of noise during the data annotation or collection phase, which we annotated as \(\epsilon_i\) and  \(\epsilon_j\), resulting in sample $i$'s ground truth label \( y_i \) being altered to \( y^*_{i} =y_i + \epsilon_i \) and sample $j$'s label \( y_j \) being altered to \(y^*_{j}= y_j + \epsilon_j \). We then obtain a noise-considering formula of delta label of sample pairs:
\begin{align}
\Delta y^*_{ij} &= y^*_{i} -  y^*_{j} \nonumber \\
  &= (y_i + \epsilon_i) - (y_j + \epsilon_j) \nonumber \\
  &= (y_i - y_j) - ( \epsilon_i + \epsilon_j) \nonumber \\
  &= \Delta y_{ij} - \sum \epsilon_{ij},\label{eq_6}
\end{align}
where \(\sum \epsilon_{ij}\) represents the sum of \(\epsilon_i\) and \(\epsilon_j\). Note that the ground truth label of the data is theoretically impossible to obtain because of the aleatoric uncertainty; therefore, it is evident that operation \ref{eq_6} will cause error accumulation, making it more difficult to learn. As a consequence, if we continue to use the original form of \(L_{prl}\), take \emph{MAE} type as an example without loss of generality, we obtain:
\begin{align}
   L_{prl-{noise}} &=\frac{1}{D} \sum_{i=1}^{B}\sum_{j=1}^{B}M_{ij} |\Delta s_{ij} - \Delta y^*_{ij}| \nonumber \\
   &= \frac{1}{D} \sum_{i=1}^{B}\sum_{j=1}^{B}M_{ij} |\Delta s_{ij} - \Delta y_{ij} + \sum \epsilon_{ij}|.
\end{align}
Assume the noise of the label in $D$ follows a Gaussian distribution \(Z \sim \mathcal{N}(\mu, \sigma^2)\), it is obvious that the accumulated error $(\epsilon_i\ + \epsilon_j)$ can also be viewed as a random variable following another Gaussian distribution with higher variance.

Therefore, the training process of minimizing \( L_{prl-{noise}}\) via gradient descent could be highly unstable because of the randomness induced by distribution sampling. Moreover, in some extreme cases where the noise becomes a dominant term, \(\Delta y_{ij}\) and \(\Delta y^*_{ij}\) could possibly have opposite signs. In that case, the original pairwise regression loss will force the model to predict \(\Delta s_{ij}\) as close as possible to \(\Delta y^*_{ij}\) rather than \(\Delta y_{ij}\), causing the model to optimize towards a totally opposite direction.

Our proposed AdaPRL method attempts to alleviate the problem of noisy samples by measuring the uncertainty level of each data point. In AdaPRL, we adapt the deep probabilistic model \cite{lakshminarayanan2017simple} as an auxiliary network to predict the uncertainty of each sample. Unlike traditional regression models, which only output a point estimation, the deep probabilistic model outputs a probability distribution of the data. We adapt a typical deep probabilistic model which uses the Gaussian Negative Log-Likelihood (NLL) loss function \cite{nix1994estimating} to predict the mean and variance of data, assuming that the label follows a Gaussian distribution. The loss function of the auxiliary network is:

\begin{equation}
Loss_{NLL} = \frac{1}{2\sigma^2}(y - \mu)^2 + \frac{1}{2}\log \sigma^2 .
\end{equation}

As a consequence, the deep probabilistic models can distinguish and quantify the aleatoric uncertainty in the data. Aleatoric uncertainty originates from the noise or ambiguity inherent in the data itself or the intrinsic randomness within the data. The variance \(\sigma^2\) output by the deep probabilistic model's predictive distribution can be an indicator of the level of aleatoric uncertainty while \(\mu\) is an indicator of the expectation of prediction.

We integrate a deep probabilistic model into the AdaPRL framework. As discussed above, assume that the label follows a Gaussian distribution \(Z \sim \mathcal{N}(\mu, \sigma^2)\). Firstly, we define an auxiliary deep probabilistic model with NLL loss and make the model predict the mean \(\mu_{i}\) together with the associated uncertainty \(\sigma_{i}\) for each sample \(s_{i}\). 

Secondly, for every pair of sample $(i, j)$, we measure the variation of the distribution of $(y_i - y_j)$. The distribution of $(y_i - y_j)$ will follow the following Gaussian distribution: 
\begin{equation} \label{eq9}
Z' \sim \mathcal{N}(\mu_i+\mu_j, \sigma_i^2 + \sigma_j^2 + 2\mathrm{Cov}(i,j)).
\end{equation}
Since both samples are derived independently from the dataset, for simplification, we assume that two Gaussian distributions are independent. As a result, the distribution of $(y_i - y_j)$ can be simplified to: 
\begin{equation}
Z' \sim \mathcal{N}(\mu_i+\mu_j, \sigma_i^2 + \sigma_j^2 ),
\end{equation}
because the covariance between \(y_i\) and \(y_j\) equals 0. 
 
Thirdly, we can generate an uncertainty matrix \(U\) with size $i \times j$ to represent the aleatoric uncertainty of pair $(i, j)$:
\begin{equation}
U_{ij} = \sigma_i^2 + \sigma_j^2 \label{eq_11} ,
\end{equation}
where both \(\sigma_i^2\) and \(\sigma_j^2\) are derived from the auxiliary deep probabilistic model with gradient detachment.

In the fourth step, conduct an inverse min-max transformation of matrix \(U\) to generate a normalized matrix, which is called the confidence matrix \(C\):
\begin{equation}
C_{ij} = 2\times \frac{max(U)-U_{ij}}{max(U)-min(U)}\label{eq_12},
\end{equation}
where \(max(U)\) and \(min(U)\) indicate the maximum and minimum value in matrix \(U\) respectively. An additional weight factor of 2 is included to empirically rescale the matrix. After the transformation, the value of \(C_{ij}\) is greater if the uncertainty \(U_{ij}\) is smaller and vice versa, which means that the normalized matrix \(C\) indicates the confidence of data pairs. The larger \(C_{ij}\), the more confidence we have in the joint probability distribution of data pairs $(i, j)$ and their relative difference.

Finally, combine the confidence matrix \(C\) with the original pairwise regression learning loss, then we would obtain pairwise regression loss with confidence (CPRL) :

If \emph{MAE} type pairwise learning loss is used, then pairwise regression loss becomes:
\begin{equation}
   L_{CPRL}=\frac{1}{D} \sum_{i=1}^{B}\sum_{j=1}^{B}C_{ij}M_{ij} |\Delta s_{ij} - \Delta y_{ij}|.\label{eq_13}
\end{equation}
If \emph{RMSE} type pairwise learning loss is used, then:
\begin{equation}
  L_{CPRL} = \sqrt{\frac{1}{D} \sum_{i=1}^{B}\sum_{j=1}^{B} C_{ij}M_{ij} (\Delta s_{ij}- \Delta y_{ij})^2},\label{eq_14}
\end{equation}
where $B$ is mini-batch size, and $D$ is the number of true elements of matrix M. Note that $L_{CPRL}$ are usually calculated in batches, and \( L_{CPRL}\) is set to 0 if none of the elements of the masked matrix $M$ are true.

We can obtain the final formula of our AdaPRL framework by combining $L_{CPRL}$ with any regression loss function with a fraction of \(\alpha\):
\begin{equation}
L_{AdaPRL}= L_{reg} +\alpha \cdot L_{CPRL},
\end{equation}
where \(L_{reg}\) could be any arbitrary regression loss function such as mean squared error (MSE), mean absolute error (MAE), or Huber loss. Hyperparameter \(\alpha\) is a float to control the weight of pairwise regression loss with confidence. The complete AdaPRL algorithm is described in \ref{alg:adaprl_algorithm}.

\begin{algorithm}[tb]
   \small
   \caption{AdaPRL} 
   \label{alg:adaprl_algorithm}
\begin{algorithmic}
\STATE {\bfseries Input:} Training data $\{(x_m, y_m)\}_{m=1}^N$, learning rate $\eta$, number of epochs $E$, mini-batch size $B$,pairwise loss type $type$, loss weight $\alpha$
\STATE Initialize model parameters $\theta_1$
\STATE Initialize auxiliary deep probabilistic network $\theta_2$ 
\FOR{epoch = $1$ {\bfseries to} $E$}
   \STATE Shuffle training data
   \FOR{each minibatch $\{(x_j, y_j)\}_{j=1}^B$}
       \STATE Forward point-wise prediction $\hat{y}_j = f(x_j; \theta_1)$
       \STATE Forward predictive distribution $p(y_j|x_j; \theta_2)$ and $\sigma_j$
       \STATE Compute $L_{reg}$ (take MSE loss as example):
       \STATE \quad $L_{reg} = \frac{1}{B} \sum\limits_{j=1}^B (\hat{y}_j - y_j)^2$
       \STATE Compute $\Delta s_{ij}$ and $\Delta y_{ij}$ for every pairs within batch
       \STATE Compute matrix $M$ with equation \ref{eq_2}
       \STATE Compute matrix $U$ and $C$ with equation \ref{eq_11} and \ref{eq_12}
       
       \STATE Compute $L_{CPRL}$:
       \IF{$type = \text{\emph{MAE}}$ }
       \STATE \quad $L_{CPRL}=\frac{1}{D} \sum_{i=1}^{B}\sum_{j=1}^{B}C_{ij}M_{ij} |\Delta s_{ij} - \Delta y_{ij}|$
       \ENDIF
       \IF{$type = \text{\emph{RMSE}}$ }
       \STATE \quad $ L_{CPRL} = \sqrt{\frac{1}{D} \sum_{i=1}^{B}\sum_{j=1}^{B} C_{ij}M_{ij} (\Delta s_{ij}- \Delta y_{ij})^2}$
       \ENDIF
       \STATE Combine MSE and CPRL obtain AdaPRL loss:
       \STATE \quad  $\mathcal{L}(\theta_1) = L_{reg} + \alpha L_{CPRL}$
       \STATE Compute NLL loss:
       \STATE \quad $\mathcal{L}(\theta_2) = -\frac{1}{B} \sum\limits_{j=1}^B \log p(y_j|x_j; \theta)$
       \STATE Back propagate model's gradients $\nabla_{\theta_1} \mathcal{L}(\theta_1)$
       \STATE Update parameters:
       \STATE \quad $\theta_1 \leftarrow \theta_1 - \eta \nabla_{\theta_1} \mathcal{L}(\theta_1)$
       \STATE Back propagate auxiliary model's gradients $\nabla_{\theta_2} \mathcal{L}(\theta_2)$
       \STATE Update parameters:
       \STATE \quad $\theta_2 \leftarrow \theta_2 - \eta \nabla_{\theta_2} \mathcal{L}(\theta_2)$       
   \ENDFOR
\ENDFOR
\STATE {\bfseries Output:} Trained model parameters $\theta_1$
\end{algorithmic}
\end{algorithm}

\subsection{Extending AdaPRL to the realm of multi-task learning and multivariate time series forecasting}
\subsubsection{AdaPRL for multi-task learning}
Multi-task learning, also known as multivariate learning, is an important research field in deep learning. In recent years, multi-task learning has been successfully applied to many real-world applications such as recommendation systems, time series prediction, computer vision, and other research fields \cite{ma2018entire,ma2018modeling,tang2020progressive,wen2020entire}. Thus, it is essential to extend AdaPRL from single output prediction to multi-task learning. A straightforward and naive approach is to calculate all possible pairs across all targets. Taking the mean absolute pairwise error type of $L_{CPRL}$ as an example, formally we have the naive version of multi-tasked confidence pairwise regression learning loss:

\begin{equation}
   L_{MCPRL_{nav}} =\frac{1}{D} \sum_{i=1}^{B*N}\sum_{j=1}^{B*N}C_{ij}M_{ij} |\Delta s_{ij} - \Delta y_{ij}|.\label{eq_16}
\end{equation}

Obviously, there are several problems with \(L_{MCPRL_{nav}}\). The loop over batch and targets will lead to a significant increase in computational complexity and memory usage, especially when \(N\) is relatively large. More importantly, the data type, endogenous meaning, as well as magnitude between targets could be vastly different therefore not suitable for calculating differences.

To solve this issue, we propose an improved version of $L_{CPRL}$ for multi-output networks. Instead of considering across all targets over every sample, we only focus on pairs within each target space. Assume the number of tasks is \(N\), we define a series of matrices $\{C\}_{}^k$ and $\{M\}_{}^k$. The elements \((i,j)\) of $k$\text{th} matrix \(M\) corresponding to whether sample \(i\) and sample \(j\) are a valid ranked pair on target \(k\), and elements \((i,j)\) of $k$\text{th} matrix \(C\) indicate the level of confidence in the certainty of the sample pair \(i\) and \(j\) on target \(k\).  Formally, we obtain:
\begin{equation}
   L_{MCPRL} =\frac{1}{D} \sum_{k=1}^{N}\sum_{i=1}^{B}\sum_{j=1}^{B}C^{k}_{ij}M^{k}_{ij} |\Delta s_{ijk} - \Delta y_{ijk}|.
\end{equation}

Since only the sample pairs within each target are considered, it is evident that the computational complexity reduces from \( O\left( (B N)^2 \right) \) to \( O\left( B^2 N \right) \). Therefore, the approach not only ensures that comparisons of data pairs are meaningful but also significantly improves computational efficiency.

\subsubsection{AdaPRL for multivariate time series forecasting}\label{SCPRL}
Time series forecasting is a crucial task in various domains, and deep learning models have significantly advanced research in this area. Multivariate time series forecasting is a major subtopic in time series forecasting that involves predicting future values of multiple interrelated time-dependent variables simultaneously, emphasizing the importance of analyzing and modeling the relationships between several targets that change over time. 

Similar to our analysis of multi-task learning above, to integrate AdaPRL in the multivariate time series forecasting, the naive method that compares all targets over all prediction lengths in every possible pair would be computationally expensive and memory-intensive. Instead, we focus on the inner correlation of every target over the time dimension:
\begin{equation}
   L_{MTCPRL} =\frac{1}{D} \sum_{k=1}^{N}\sum_{i=1}^{B*T}\sum_{j=1}^{B*T}C^{k}_{ij}M^{k}_{ij} |\Delta s_{ijk} - \Delta y_{ijk}|.
\end{equation}
The computational complexity reduces from \( O\left( (B N T)^2 \right) \) to \( O\left( (BT)^2 N \right) \) compared to naive pairwise learning applied to multivariate time series forecasting. Since the number of targets in many multivariate datasets could be relatively large, this adoption could save computing and memory and work well with short-term time series forecasting.

However, in the context of long-term time series forecasting, the additional computational complexity remains significant because the prediction sequence could be \(720\) or even longer \cite{oreshkin_n-beats_2019,nie_time_2022,zhang2023crossformer,zeng2023transformers}. To address this challenge, we further propose a novel \textbf{Sparse Confidence Pairwise Learning (SCPRL)}, annotated as \(L_{SCPRL}\). The definition is:
\begin{equation}
   L_{SCPRL} =\frac{1}{D} \sum_{k=1}^{N}\sum_{(i,j) \in S}C^{k}_{ij}M^{k}_{ij} |\Delta s_{ijk} - \Delta y_{ijk}|,
\end{equation}
where \(S\) indicates a sparse area that can be derived either randomly or heuristically. Note that the area of \(S\) satisfies \(|S|\ll B^2T^2\); therefore, the computational complexity and memory consumption reduce significantly from \( O\left( (BT)^2 N \right)\) to \( O\left( |S| N \right)\). 

Furthermore, if randomness is induced when generating the sparse area \(S\), \(L_{SCPRL}\) can also be viewed as a variant of dropout \cite{srivastava2014dropout} in Confidence Pairwise Regression Loss of \ref{eq_13}. Conclusively, our sparse confidence pairwise learning approach only conducts comparisons within a certain area instead of all pairs, takes the most advantage of learning patterns over a longer time horizon while theoretically reducing overfitting and maintaining computation and memory efficiency. Note that sparse confidence pairwise learning loss can be integrated with not just multivariate time series forecasting frameworks, but also single-output, multi-task learning and other sophisticated models simply by adjusting the definition of sparse area \(S\).

\begin{table*}[]
\vskip 0.1in
\begin{adjustbox}{width=\linewidth,center}
\setlength{\tabcolsep}{4pt}
\centering
{\large
{
 \renewcommand{\arraystretch}{1.2}
\begin{sc}
\begin{tabular}{@{}*{3}{c}|*{3}{c}|*{3}{c}|*{2}{c}|*{3}{c}|*{3}{c}|*{3}{c}|*{1}{c}|*{2}{c}@{}}
\toprule
\multicolumn{1}{c}{} && \multicolumn{21}{c}{Evaluation Dataset (Model) } \\
\cmidrule(l){3-23}
\multicolumn{1}{c}{} & \multicolumn{1}{c}{} && \multicolumn{3}{c}{Movielens-1M (MLP)} & \multicolumn{3}{c}{KuaiRec (MLP)} & \multicolumn{2}{c}{ETT (Minusformer)}& \multicolumn{3}{c}{AgeDB (ResNet50)} & \multicolumn{3}{c}{CLAP2016 (ResNet50)} & \multicolumn{3}{c}{Jigsaw (RoBERTa)} & \multicolumn{1}{c}{Janestreet (MHA)} & \multicolumn{2}{c}{Industrial (MLP)} \\
\cmidrule(l){4-6}
\cmidrule(l){7-9}
\cmidrule(l){10-11}
\cmidrule(l){12-14}
\cmidrule(l){15-17}
\cmidrule(l){18-20}
\cmidrule(l){21-21}
\cmidrule(l){22-23} 

 \multicolumn{3}{r}{\diagbox[height=0.7cm]{Method\quad}{Metric}} & \multicolumn{1}{c}{MSE $\downarrow$} & \multicolumn{1}{c}{MAE$ \downarrow$} & \multicolumn{1}{c}{Kendall's \(\tau\) $\uparrow$} & \multicolumn{1}{c}{MSE $\downarrow$} & \multicolumn{1}{c}{MAE $\downarrow$} & \multicolumn{1}{c}{Kendall's \(\tau\) $\uparrow$}& \multicolumn{1}{c}{MSE $\downarrow$} & \multicolumn{1}{c}{MAE $\downarrow$} &  \multicolumn{1}{c}{MSE $\downarrow$} & \multicolumn{1}{c}{MAE $\downarrow$} & \multicolumn{1}{c}{Kendall's \(\tau\) $\uparrow$} & \multicolumn{1}{c}{MSE $\downarrow$} & \multicolumn{1}{c}{MAE $\downarrow$} & \multicolumn{1}{c}{Kendall's \(\tau\) $\uparrow$} & \multicolumn{1}{c}{MSE $\downarrow$} & \multicolumn{1}{c}{MAE $\downarrow$} & \multicolumn{1}{c}{Kendall's \(\tau\) $\uparrow$} &  \multicolumn{1}{c}{Weighted $R^2$ $\uparrow$} & \multicolumn{1}{c}{MSE $\downarrow$} & \multicolumn{1}{c}{MAE $\downarrow$} \\

\cmidrule(l){3-23} 
&& \multicolumn{1}{l}{Baseline (L2)} & 0.7418 & 0.6787 & 0.4873 & 0.9438 & 0.2859 & 	0.5671 & 0.3677 & 0.3868 & 69.939 & 6.372 & 0.697 & 45.824 & 4.726 & 0.753  & 0.3686 & 0.3010 & 0.6843 & 0.0540 & 0.5301 & 0.2941 \\
\cmidrule(l){3-23}

&& \multicolumn{1}{l}{ RankSim\cite{gong2022ranksim}} & 0.7473 & 0.6818 & 0.4833 & \textbf{0.9338} & \underline{0.2700} & \underline{0.5761} & - & - & 68.938 & 6.368& 0.698 & 43.161 & 4.547 & \textbf{0.763} & \textbf{0.3440} & 0.3041& \underline{0.6872} & - & 0.5288 & 0.2927   \\
&& \multicolumn{1}{l}{ Ordinal Entropy\cite{zhang2023improving}} & 0.7424 & 0.6790 & 0.4868 & 1.1825&0.4122&0.4699 & - & - &\underline{68.873} & 6.380 &\underline{0.699} & \underline{42.549}&\underline{4.535}&0.757 & 0.3626& \textbf{0.2892} & \textbf{0.6897} & - & \underline{0.5269} & 0.2899  \\

\cmidrule(l){3-23} 
&& \multicolumn{1}{l}{Pairwise Regression Learning} & \underline{0.7398} & \underline{0.6782} & \underline{0.4889} & 0.9368& 0.2797 & 0.5679 & \underline{0.3669} & \underline{0.3855} & \textbf{68.559} & \underline{6.353} & 0.698  & 44.377 & 4.545 & 0.757  & 0.3569 & 0.2988 & 0.6858 & \underline{0.0575}  & 0.5274 & \underline{0.2894} \\
&& \multicolumn{1}{l}{AdaPRL} & \textbf{0.7384} & \textbf{0.6764} & \textbf{0.4895} & \underline{0.9360}&\textbf{0.2661}&\textbf{0.5844}  
 & \textbf{0.3634} & \textbf{0.3810} & 69.342& \textbf{6.327} & \textbf{0.699} & \textbf{42.397} & \textbf{4.531} & \underline{0.759}  & \underline{0.3561} & \underline{0.2922}& 0.6858 & \textbf{0.0617} & \textbf{0.5250} & \textbf{0.2838}   \\
\cmidrule(l){3-23}
&& \multicolumn{1}{l}{\emph{Rel.Imp}} & 0.46\% &0.34\% &0.45\% & 0.83\% & 6.93\% & 3.05\% & 1.19\% & 1.52\% & 0.85\% & 0.71\% & 0.29\%
 & 7.48\%&4.13\%&0.80\% & 3.39\% & 2.92\% & 0.22\% & 14.25\% & 0.96\% & 1.93\%     \\
\bottomrule
\end{tabular}
\end{sc}
}}
\end{adjustbox}
\caption{Performance comparison of all methods across regression datasets concerning Mean Squared Error (MSE), Mean Absolute Error (MAE), and Kendall's \(\tau\). We bold the best performance, while the underlined scores were the second best of each dataset. \emph{Rel.Imp} indicates the relative improvement of AdaPRL over the baseline. The results of RankSim and Ordinal Entropy for the ETT and Janestreet datasets were not reported because they were not compatible with the time series forecasting task. Detailed performance of each method on individual datasets was elaborated in the experiment section of the Appendix.}
\vskip -0.1in
\label{overall-table}
\label{tab:large_table}
\end{table*}

\section{Experiments}
In this section, we conducted extensive experiments on a variety of regression datasets in different fields for the purpose of demonstrating that this method had universal generalization capabilities not only across different datasets but also across different backbone networks. Datasets included tabular regression datasets, natural language, and vision datasets. The research field of data also varied from recommendation, time-series prediction, age prediction, natural language understanding, finance, etc. Moreover, we conducted offline evaluations and online AB tests with millions of users, which proved that our method also benefited real-world industrial applications. Due to page limitations, we presented only the overall results for representative datasets in the main body of the paper. Comprehensive experimental details were thoroughly discussed in the Appendix \ref{app_sec_a}.

\subsection{Experiment Setup}
To comprehensively evaluate the proposed method, an extensive experiment was conducted to assess the performance across multiple real-world datasets. The data domains included recommendation systems, time series forecasting, age prediction, natural language understanding, finance, and industrial data. Specifically, the evaluation datasets included Movielens-1M, KuaiRec, ETT, AgeDB, CLAP2016, Jigsaw, Janestreet, and a large-scale proprietary industrial dataset from a major music streaming app. The statistics of the datasets are summarized in Table \ref{data-stats-tbl}. We chose MSE loss as the baseline method, denoted as Baseline(L2), and we chose RankSim \cite{gong2022ranksim} and Ordinal Entropy \cite{zhang2023improving} for comparison because they were claimed to be effective in regression tasks. We also adapted the original form of pairwise regression loss without uncertainty estimation described in equations \ref{eq_4} and \ref{eq_5} for comparison. Please refer to the Appendix \ref{app_sec_a} for detailed experiment setups as well as the experiment results for all datasets.

\begin{table}[t]
\vskip 0.15in
\begin{center}
\begin{small}
\begin{sc}\resizebox{1.0\columnwidth}{!}{
\begin{tabular}{lcccc}
\toprule
Domain & Data & \#Samples  & \#fields & \#Features  \\ 
\midrule
\multirow{2}{*}[-0.5ex]{\textbf{Recommend. system}} & Movielens-1M & 1,000,209 & 7 & 10,082  \\ 
 & KuaiRec(subset) & 7,051,542 & 85 & 5,084  \\ 
 \midrule
 \multirow{2}{*}[-0.5ex]{\textbf{Time series forecast.}} & ETT & 174,204 & 7 & 5,040  \\ 
  & Electricity & 26,305 & 321 & 231,120  \\ 
 \midrule
 \multirow{2}{*}[-0.5ex]{\textbf{Age prediction}} & AgeDB & 16,488 & - & 65,536  \\ 
  & CLAP2016 & 7,591 & - & 65,536  \\ 
 \midrule
  \multirow{1}{*}[-0.5ex]{\textbf{NLU}} & Jigsaw & 47,792 & - & 128  \\ 
 \midrule
  \multirow{1}{*}[-0.5ex]{\textbf{Financial}} & Janestreet & 47,127,338 & 79 & 5,056  \\ 
\midrule
  \multirow{1}{*}[-0.5ex]{\textbf{Industrial}} & Music App & 49,164,901 & 445 & 1,241,513  \\ 
\bottomrule
\end{tabular}
}
\end{sc}
\end{small}
\caption{Statistics of datasets used in the experiments section. Note that \emph{\#Features} means differently in different domain's dataset. For recommendation data it indicates the number of sparse features, while for image datasets and nature language data, it indicates the number of pixels and number of tokens of each sample respectively, while \emph{\#Fields} denotes number of individual columns for tabular data and denotes number of task in time series forecasting dataset.}
\label{data-stats-tbl}
\end{center}
\vskip -0.1in
\end{table}

\subsection{Overall Performance}
As shown in Table \ref{overall-table}, we compared several commonly used regression loss functions on various datasets of multiple domains, including the Mean Squared Error (MSE) (also known as the $L2$) loss, as well as regression learning methods such as RankSim \cite{gong2022ranksim} and Ordinal Entropy \cite{zhang2023improving} method, alongside our proposed methods pairwise regression learning (PRL) and AdaPRL. We conducted a comprehensive horizontal comparison of the key performance of these regression methods across various datasets, including Movielens-1M, KuaiRec, ETT, ECL, AgeDB, CLAP2016, Jigsaw, Janestreet, and an industrial dataset. As indicated in the table, our proposed method AdaPRL achieved the best performance on the vast majority of datasets, surpassing the previously introduced regression learning methods RankSim and Ordinal Entropy in all datasets except KuaiRec, thereby demonstrating significant potential and generalization capabilities. In addition, the pairwise regression learning (PRL) method also exhibited a certain degree of improvement over the original MSE baseline on most datasets. Furthermore, Kendall's \(\tau\) of our proposed method was also higher compared with traditional loss functions, indicating that it further improves ranking ability, which helps distinguish the monotonicity of data.

\subsection{Online A/B Testing}

To evaluate the effectiveness of the AdaPRL algorithm in real-world commercial applications, we deployed it with tuned hyperparameters in a music streaming application that serves tens of millions of active users every day. The regression task is a crucial component of the value estimation module within our user-targeting framework, which is integral to a commercial business generating millions in daily revenue; therefore, even minor improvements in the accuracy can yield significant commercial benefits. Specifically, we conducted an online A/B test comprising a control group using the original mean squared error loss and an experimental group applying the AdaPRL algorithm in the regression component, each representing 10\% of our total traffic of the App. Over a two-week period, we measured the actual revenue per user generated by both groups. As shown in Table \ref{qm-table-abt}, the average daily income per user resulting from the use of the AdaPRL algorithm in regression increased by 2.57\% compared to the same model using the MSE loss, and the increase was statistically significant according to the t-test with \textit{p}-value $<$ 0.05, suggesting that AdaPRL could enhance commercial benefits in real-world applications.

\begin{table}[t]
\vskip 0.15in
\begin{center}
\begin{small}
\begin{sc}\resizebox{1.0\columnwidth}{!}{
\begin{tabular}{ccc}
\toprule
Online A/B testing Time & Rel.Imp. of revenue per user & \textit{p}-value \\
\midrule
14 days & 2.57\% & $<$ 0.05 \\ 
\bottomrule
\end{tabular}
}
\end{sc}
\end{small}
\caption{Online A/B testing results for MSE and AdaPRL on the large-scale proprietary industrial regression component of a major music streaming App over a two-week period.}

\label{qm-table-abt}
\end{center}
\vskip -0.1in
\end{table}

\subsection{Further Study and Analysis}
To facilitate a deeper understanding of the underlying mechanisms, such as the sensitivity of hyperparameters,the effect of sparsity, we conducted several comprehensive experiments for our proposed method. Additionally, various experimental analyses were performed to examine the model's robustness and generalizability across different scenarios and datasets, which demonstrates that AdaPRL improves the model's robustness to label noise in the training data, resilience to unforeseen data corruption, and robustness to reduced training data. Moreover, extensive experiments show that AdaPRL can be seamlessly incorporated into various regression frameworks, such as RankSim and Ordinal Entropy, and boost their performance. In addition, we visualize the results on the time series forecasting task to demonstrate that AdaPRL not only enhances predictive accuracy but also provides confidence intervals for predictions through uncertainty quantification, thereby improving the interpretability of the results. Theoretical and experimental analysis shows that the extra computational cost of AdaPRL is generally considered acceptable. Due to the limited page, the detailed experimental settings, results, discussions, and further study are relegated to the Appendix.

\section{Conclusion}
In this paper, we propose a novel pairwise regression learning framework called AdaPRL, which leverages the relative differences between data points and integrates with deep probabilistic models to quantify the uncertainty associated with the predictions. Extensive experiments, as well as online A/B testing in a major music streaming App, demonstrate that our method has the potential to enhance the accuracy, robustness, and interpretability of modern deep learning regression models across various domains, including recommendation systems, time series forecasting, age prediction, natural language processing, finance, and real-world industrial applications.

\section*{Impact Statement}
This paper focuses on proposing a generalized and robust pairwise learning algorithm to enhance the performance of universal regression tasks in various domains of deep learning, and push the accuracy limit to a higher level. We believe that this work is helpful to advance the goal. There are many potential societal consequences of our work, none of which we feel must be specifically highlighted here.

\newpage
\nocite{langley00}

\bibliography{adaprl_arxiv}
\bibliographystyle{arxiv}

\clearpage

\appendix

\section{Experiments Details}\label{app_sec_a}
\subsection{Recommendation Systems}
\subsubsection{Datasets}
\textbf{MovieLens-1M\footnote{https://grouplens.org/datasets/movielens/1m/}} \cite{harper2015movielens} is a popular dataset in recommendation systems research that contains 1 million records of the (user-features, item-features, rating) triplet, where the rating scores are between 1 and 5. Similar to previous work \cite{song_autoint_2019,wang_dcn_2020}, we also formalize the task as a rating regression problem. We use all single-value categorical features as the input of the embedding layer, followed by a multilayer perceptron to predict the target. The data is randomly split into 80\% for training, 10\% for validation, and 10\% for testing.

\textbf{KuaiRec\footnote{https://kuairec.com/}} \cite{gao_kuairec_2022}  is a real-world dataset collected from the recommendation logs of a famous mobile video sharing app with a fully observed user-item interaction matrix that originally aims to be used in unbiased offline evaluation for recommendation models. We employ the fully observed small matrix for both training, validation, and testing. All user and item features are categorized using feature binning. These features are then processed through an embedding layer, followed by a multilayer perceptron, to predict video completion rates. The data is randomly split into 80\% for training, 10\% for validation, and 10\% for testing.

\subsubsection{Experiment Setup}
We performed an early stopping strategy with a patience of three in the valid set using MSE metrics. Every experiment was repeated five times and the average of corresponding metrics in terms of mean squared error (MSE), mean absolute error (MAE), and Kendall's \(\tau\) was reported. To demonstrate the robustness of our method to hyperparameters, we only performed a coarse-grained grid search in most of our experiments. The hyperparameters include pairwise ranking loss type (MAE or RMSE type as described in 3.2.2) and the weight factor \(\alpha\) between \{0.01,0.02,0.05,0.1,0.2,0.5\}.
 
\begin{table}[t]
\vskip 0.15in
\begin{center}
\begin{small}
\begin{sc}\resizebox{1.0\columnwidth}{!}{
\begin{tabular}{lcccc}
\toprule
Dataset & loss type & MSE$ \downarrow$ & MAE$ \downarrow$ & Kendall‘s \(\tau\) $\uparrow$\\
\midrule

\multirow{15}{*}[-1.5ex]{\textbf{ML-1M}}   
  & L1 & 0.9139&0.6957&0.4153  \\ 
  & L1+PRL & \underline{0.7908} & \underline{0.6674} & \underline{0.4741} \\
  & AdaPRL(L1) & \textbf{0.7785} & \textbf{0.6627} & \textbf{0.4838} \\
 \cmidrule(lr){2-5}
& \emph{Rel.Imp} & 14.82\% & 4.74\% & 16.49\% \\
  \cmidrule(lr){2-5}
  & Huber & 0.7592 & 0.6836 & 0.4798 \\
  &  Huber+PRL & \underline{0.7447} & \underline{0.6755} & \underline{0.4882} \\
  & AdaPRL(Huber) & \textbf{0.7436} & \textbf{0.6738} & \textbf{0.4898} \\
 \cmidrule(lr){2-5}
& \emph{Rel.Imp} & 2.05\% & 1.43\% & 2.08\% \\
  \cmidrule(lr){2-5}
  \cmidrule(lr){2-5}
  & L2 &0.7418 &0.6787 &	0.4873  \\ 
  & L2+PRL & \underline{0.7398} & \underline{0.6782} & \underline{0.4889 } \\
  & AdaPRL(L2) & \textbf{0.7384} & \textbf{0.6764} & \textbf{0.4895} \\
  \cmidrule(lr){2-5}
  & \emph{Rel.Imp} & 0.46\% & 0.34\% & 0.45\% \\
\bottomrule
\end{tabular}
}
\end{sc}
\end{small}
\caption{Regression accuracies for Pairwise Regression Learning and AdaPRL on ML-1M. \emph{Rel.Imp} indicates relative improvement over corresponding metrics. The experiments were repeated five times, and the average results were reported.}
\label{ml-1m-table}
\end{center}
\vskip -0.1in
\end{table}

\begin{table}[t]
\vskip 0.15in
\begin{center}
\begin{small}
\begin{sc}\resizebox{1.0\columnwidth}{!}{
\begin{tabular}{lcccc}
\toprule
Dataset & loss type & MSE$\downarrow$ & MAE$\downarrow$ & Kendall‘s \(\tau\)$\uparrow$ \\
\midrule

\multirow{15}{*}[-1.5ex]{\textbf{KuaiRec}}   
  & L1 & 1.0334&0.2707&0.6018 \\ 
  & L1+PRL & \underline{0.9905}& \underline{0.2562} & \underline{0.6049}\\
  & AdaPRL(L1) & \textbf{0.9816}& \textbf{0.2524} & \textbf{0.6067}\\
 \cmidrule(lr){2-5}
& \emph{Rel.Imp} & 5.01\% & 6.76\% & 0.81\%\\
  \cmidrule(lr){2-5}
  & Huber & 0.9522& 0.2648 & 0.5976 \\
  &  Huber+PRL & \underline{0.9463}& \underline{0.2586} & \underline{0.5980}\\
  & AdaPRL(Huber) & \textbf{0.9378}& \textbf{0.2584} & \textbf{0.6020}\\
 \cmidrule(lr){2-5}
& \emph{Rel.Imp} & 1.51\% & 2.42\% & 0.74\%\\
  \cmidrule(lr){2-5}
  \cmidrule(lr){2-5}
  & L2 & 0.9438& 0.2859 & 0.5671 \\ 
  & L2+PRL & \underline{0.9368}& \underline{0.2797} & \underline{0.5679}\\
  & AdaPRL(L2) & \textbf{0.9360}& \textbf{0.2661} & \textbf{0.5844}\\
  \cmidrule(lr){2-5}
  & \emph{Rel.Imp} & 0.83\% & 6.93\% & 3.05\%\\
\bottomrule
\end{tabular}
}
\end{sc}
\end{small}
\caption{Regression accuracies for Pairwise Regression Learning and AdaPRL on KuaiRec. \emph{Rel.Imp} indicates relative improvement over corresponding metrics. The experiments were repeated five times, and the average results were reported.}
\label{kuairec-table}
\end{center}
\vskip -0.1in
\end{table}

\subsubsection{Overall Performance}
We categorized the dataset and bolded the best performance, while the underlined scores were the second best. The experimental results are shown in Table \ref{ml-1m-table} and \ref{kuairec-table}. We can draw the following conclusions: by comparing using single \(L2\) loss, \(L2\) with original pairwise learning (\(L2+PRL\)) and our AdaPRL method, the AdaPRL achieved the highest accuracy among all other methods in Movielens-1M. Specifically, the test MSE and MAE metrics were reduced by 14.82\% and 4.74\%, 2.05\% and 1.43\%, and 0.46\% and 0.34\% respectively with \(L1\), \(Huber\) and \(L2\). AdaPRL also achieved the best performance in the KuaiRec dataset when combined with three different \(L_{reg}\). The test MSE and MAE metrics were reduced by 5.01\% and 6.76\%, 1.51\% and 2.42\%, and 0.83\% and 6.93\% respectively with \(L1\), \(Huber\) and \(L2\). Moreover, the Kendall's \(\tau\) was increased by 0.45\% and 3.05\% respectively, on the ML-1M and KuaiRec datasets with \(L2\), indicating that the model not only generates more accurate predictions but also provides a stronger monotonic relationship between labels and predictions.

\subsection{Time Series Forecasting}
\subsubsection{Datasets}
\textbf{The Electricity Transformer Temperature (ETT) dataset\footnote{https://github.com/zhouhaoyi/ETDataset}} \cite{haoyietal-informer-2021} is a multivariate time-series forecasting dataset comprising two years of data from two distinct regions. ETT is a critical indicator for long-term electric power deployment. The dataset is divided into four subsets: ETTh1 and ETTh2, which are at the 1-hour level, and ETTm1 and ETTm2, which are at the 15-minute level. Each data point includes the target value, oil temperature, along with six features of the power load. The dataset is divided into training, validation, and test sets of 12, 4, and 4 months, respectively.

\textbf{Electricity dataset\footnote{https://github.com/laiguokun/multivariate-time-series-data}} \cite{wu2021autoformer}, also known as the ECL dataset, is another multivariate time series forecasting dataset. The ECL data contains electricity consumption information from 2012 to 2014, and it records the hourly electricity consumption data of 321 clients.

\subsubsection{Experiment Setup}
\label{ts-exp-setup}
Following recent works \cite{oreshkin_n-beats_2019,nie_time_2022,zhang2023crossformer,zeng2023transformers,haoyietal-informer-2021,liu_itransformer_2024}, we conducted our experiments on four subsets of ETT as well as the Electricity dataset. The input length was fixed to 96 as the previous work did, and the prediction lengths were between \{96,192,336,720\} respectively. Note that a randomized sparse version of confident pairwise regression learning was used when the prediction length was 720 for ETT and 192/336/720 for Electricity. We adopted two state-of-the-art time series models, including iTransformer \cite{liu_itransformer_2024} and Minusformer \cite{liang_minusformer_2024} as strong baselines to show the generalizability of our proposed AdaPRL method. The pairwise loss type was set to MAE to reduce penalty over outliers, and a coarse-grained grid search was performed on a weight factor \(\alpha\) between \{0.5,1.0,2.0,4.0,8.0,16.0,32.0,64.0\}.

\renewcommand\arraystretch{1.2}
\begin{table}[t]

\vskip 0.15in
\begin{center}
\begin{small}
\begin{sc}
\resizebox{1.0\columnwidth}{!}{
\begin{tabular}{lcccccc}
\toprule
Model & \multicolumn{2}{|l|}{iTransformer} & \multicolumn{2}{|l|}{iTrans.+AdaPRL} & \multicolumn{2}{|l|}{Relative Improv.} \\
\toprule
Dataset & \multicolumn{1}{|c|}{MSE$\downarrow$} & \multicolumn{1}{|c|}{MAE$\downarrow$} & \multicolumn{1}{|c|}{MSE$\downarrow$} & \multicolumn{1}{|c|}{MAE$\downarrow$}& \multicolumn{1}{|c|}{MSE$\downarrow$} & \multicolumn{1}{|c|}{MAE$\downarrow$}  \\
\toprule
ETTM1\_96\_96  & \multicolumn{1}{|c|}{0.334} & \multicolumn{1}{|c|}{0.368} & \multicolumn{1}{|c|}{\textbf{0.327}} & \multicolumn{1}{|c|}{\textbf{0.359}} & \multicolumn{1}{|c|}{2.03\%} & \multicolumn{1}{|c|}{2.51\%}\\ 
ETTM1\_96\_192  & \multicolumn{1}{|c|}{0.377} & \multicolumn{1}{|c|}{0.391} & \multicolumn{1}{|c|}{\textbf{0.375}} & \multicolumn{1}{|c|}{\textbf{0.383}} & \multicolumn{1}{|c|}{0.66\%} & \multicolumn{1}{|c|}{2.07\%}\\
ETTM1\_96\_336  & \multicolumn{1}{|c|}{0.426} & \multicolumn{1}{|c|}{0.420} & \multicolumn{1}{|c|}{\textbf{0.410}} & \multicolumn{1}{|c|}{\textbf{0.409}} & \multicolumn{1}{|c|}{3.81\%} & \multicolumn{1}{|c|}{2.82\%}\\
ETTM1\_96\_720\textsuperscript{†}  & \multicolumn{1}{|c|}{0.491} & \multicolumn{1}{|c|}{0.459} & \multicolumn{1}{|c|}{\textbf{0.481}} & \multicolumn{1}{|c|}{\textbf{0.449}} & \multicolumn{1}{|c|}{2.03\%} & \multicolumn{1}{|c|}{2.22\%}\\
\midrule
ETTM2\_96\_96  & \multicolumn{1}{|c|}{0.180} & \multicolumn{1}{|c|}{0.264} & \multicolumn{1}{|c|}{\textbf{0.178}} & \multicolumn{1}{|c|}{\textbf{0.257}} & \multicolumn{1}{|c|}{1.02\%} & \multicolumn{1}{|c|}{2.81\%}\\
ETTM2\_96\_192  & \multicolumn{1}{|c|}{0.250} & \multicolumn{1}{|c|}{0.309} & \multicolumn{1}{|c|}{\textbf{0.245}} & \multicolumn{1}{|c|}{\textbf{0.301}} & \multicolumn{1}{|c|}{1.79\%} & \multicolumn{1}{|c|}{2.46\%}\\
ETTM2\_96\_336  & \multicolumn{1}{|c|}{0.311} & \multicolumn{1}{|c|}{0.348} & \multicolumn{1}{|c|}{\textbf{0.309}} & \multicolumn{1}{|c|}{\textbf{0.343}} & \multicolumn{1}{|c|}{0.57\%} & \multicolumn{1}{|c|}{1.33\%}\\
ETTM2\_96\_720\textsuperscript{†}  & \multicolumn{1}{|c|}{0.412} & \multicolumn{1}{|c|}{0.407} & \multicolumn{1}{|c|}{\textbf{0.404}} & \multicolumn{1}{|c|}{\textbf{0.398}} & \multicolumn{1}{|c|}{2.03\%} & \multicolumn{1}{|c|}{2.23\%}\\
\midrule
ETTH1\_96\_96  & \multicolumn{1}{|c|}{0.386} & \multicolumn{1}{|c|}{0.405} & \multicolumn{1}{|c|}{\textbf{0.384}} & \multicolumn{1}{|c|}{\textbf{0.397}} & \multicolumn{1}{|c|}{0.73\%} & \multicolumn{1}{|c|}{1.49\%}\\
ETTH1\_96\_192  & \multicolumn{1}{|c|}{0.441} & \multicolumn{1}{|c|}{0.436} & \multicolumn{1}{|c|}{\textbf{0.438}} & \multicolumn{1}{|c|}{\textbf{0.431}} & \multicolumn{1}{|c|}{0.59\%} & \multicolumn{1}{|c|}{1.06\%}\\
ETTH1\_96\_336  & \multicolumn{1}{|c|}{0.487} & \multicolumn{1}{|c|}{0.458} & \multicolumn{1}{|c|}{\textbf{0.482}} & \multicolumn{1}{|c|}{\textbf{0.454}} & \multicolumn{1}{|c|}{0.97\%} & \multicolumn{1}{|c|}{0.93\%}\\
ETTH1\_96\_720\textsuperscript{†}  & \multicolumn{1}{|c|}{0.503} & \multicolumn{1}{|c|}{0.491} & \multicolumn{1}{|c|}{\textbf{0.501}} & \multicolumn{1}{|c|}{\textbf{0.489}} & \multicolumn{1}{|c|}{0.44\%} & \multicolumn{1}{|c|}{0.38\%}\\
\midrule
ETTH2\_96\_96  & \multicolumn{1}{|c|}{0.297} & \multicolumn{1}{|c|}{0.349} & \multicolumn{1}{|c|}{\textbf{0.290}} & \multicolumn{1}{|c|}{\textbf{0.339}} & \multicolumn{1}{|c|}{2.50\%} & \multicolumn{1}{|c|}{2.82\%}\\
ETTH2\_96\_192  & \multicolumn{1}{|c|}{0.380} & \multicolumn{1}{|c|}{0.400} & \multicolumn{1}{|c|}{\textbf{0.370}} & \multicolumn{1}{|c|}{\textbf{0.391}} & \multicolumn{1}{|c|}{2.84\%} & \multicolumn{1}{|c|}{2.83\%}\\
ETTH2\_96\_336  & \multicolumn{1}{|c|}{0.428} & \multicolumn{1}{|c|}{0.432} & \multicolumn{1}{|c|}{\textbf{0.416}} & \multicolumn{1}{|c|}{\textbf{0.426}} & \multicolumn{1}{|c|}{2.81\%} & \multicolumn{1}{|c|}{1.48\%}\\
ETTH2\_96\_720\textsuperscript{†}  & \multicolumn{1}{|c|}{0.427} & \multicolumn{1}{|c|}{0.445} & \multicolumn{1}{|c|}{\textbf{0.419}} & \multicolumn{1}{|c|}{\textbf{0.442}} & \multicolumn{1}{|c|}{1.84\%} & \multicolumn{1}{|c|}{0.71\%}\\
\midrule
\textbf{AVG} & \multicolumn{1}{|c|}{0.380} & \multicolumn{1}{|c|}{0.396} & \multicolumn{1}{|c|}{\textbf{0.374}} & \multicolumn{1}{|c|}{\textbf{0.389}} & \multicolumn{1}{|c|}{1.65\%} & \multicolumn{1}{|c|}{1.86\%}\\
\bottomrule
\end{tabular}
}

\end{sc}
\end{small}
\caption{Regression accuracies for AdaPRL on 4 subsets of the ETT dataset using iTransformer as the backbone model. The experiments of AdaPRL were repeated three times, and the average of the corresponding metrics was reported. Sign\textsuperscript{†} indicated using a 50\% randomized sparse confidence pairwise learning due to memory bottleneck.}
\label{itransformer-table-ETT}
\end{center}
\vskip -0.1in
\end{table}

\renewcommand\arraystretch{1.2}
\begin{table}[t]
\vskip 0.15in
\begin{center}
\begin{small}
\begin{sc}
\resizebox{1.0\columnwidth}{!}{
\begin{tabular}{lcccccc}
\toprule
Model & \multicolumn{2}{|l|}{Minusformer} & \multicolumn{2}{|l|}{Minus.+AdaPRL} & \multicolumn{2}{|l|}{Relative Improv.} \\
\toprule
Dataset & \multicolumn{1}{|c|}{MSE$\downarrow$} & \multicolumn{1}{|c|}{MAE$\downarrow$} & \multicolumn{1}{|c|}{MSE$\downarrow$} & \multicolumn{1}{|c|}{MAE$\downarrow$}& \multicolumn{1}{|c|}{MSE$\downarrow$} & \multicolumn{1}{|c|}{MAE$\downarrow$}  \\
\toprule
ETTM1\_96\_96  & \multicolumn{1}{|c|}{0.317} & \multicolumn{1}{|c|}{0.355} & \multicolumn{1}{|c|}{\textbf{0.312}} & \multicolumn{1}{|c|}{\textbf{0.346}} & \multicolumn{1}{|c|}{1.61\%} & \multicolumn{1}{|c|}{2.64\%}\\ 
ETTM1\_96\_192  & \multicolumn{1}{|c|}{0.363} & \multicolumn{1}{|c|}{0.380} & \multicolumn{1}{|c|}{\textbf{0.361}} & \multicolumn{1}{|c|}{\textbf{0.373}} & \multicolumn{1}{|c|}{0.54\%} & \multicolumn{1}{|c|}{1.78\%}\\
ETTM1\_96\_336  & \multicolumn{1}{|c|}{0.398} & \multicolumn{1}{|c|}{0.405} & \multicolumn{1}{|c|}{\textbf{0.397}} & \multicolumn{1}{|c|}{\textbf{0.400}} & \multicolumn{1}{|c|}{0.24\%} & \multicolumn{1}{|c|}{1.37\%}\\
ETTM1\_96\_720\textsuperscript{†}  & \multicolumn{1}{|c|}{0.457} & \multicolumn{1}{|c|}{0.445} & \multicolumn{1}{|c|}{\textbf{0.454}} & \multicolumn{1}{|c|}{\textbf{0.438}} & \multicolumn{1}{|c|}{0.66\%} & \multicolumn{1}{|c|}{1.53\%}\\
\midrule
ETTM2\_96\_96  & \multicolumn{1}{|c|}{0.176} & \multicolumn{1}{|c|}{0.257} & \multicolumn{1}{|c|}{\textbf{0.173}} & \multicolumn{1}{|c|}{\textbf{0.252}} & \multicolumn{1}{|c|}{1.86\%} & \multicolumn{1}{|c|}{2.18\%}\\
ETTM2\_96\_192  & \multicolumn{1}{|c|}{0.241} & \multicolumn{1}{|c|}{0.301} & \multicolumn{1}{|c|}{\textbf{0.237}} & \multicolumn{1}{|c|}{\textbf{0.295}} & \multicolumn{1}{|c|}{1.69\%} & \multicolumn{1}{|c|}{1.97\%}\\
ETTM2\_96\_336  & \multicolumn{1}{|c|}{0.302} & \multicolumn{1}{|c|}{0.340} & \multicolumn{1}{|c|}{\textbf{0.298}} & \multicolumn{1}{|c|}{\textbf{0.334}} & \multicolumn{1}{|c|}{1.10\%} & \multicolumn{1}{|c|}{1.75\%}\\
ETTM2\_96\_720\textsuperscript{†}  & \multicolumn{1}{|c|}{0.397} & \multicolumn{1}{|c|}{0.395} & \multicolumn{1}{|c|}{\textbf{0.392}} & \multicolumn{1}{|c|}{\textbf{0.390}} & \multicolumn{1}{|c|}{1.08\%} & \multicolumn{1}{|c|}{1.22\%}\\
\midrule
ETTH1\_96\_96  & \multicolumn{1}{|c|}{0.371} & \multicolumn{1}{|c|}{0.393} & \multicolumn{1}{|c|}{\textbf{0.369}} & \multicolumn{1}{|c|}{\textbf{0.388}} & \multicolumn{1}{|c|}{0.40\%} & \multicolumn{1}{|c|}{1.40\%}\\
ETTH1\_96\_192  & \multicolumn{1}{|c|}{0.432} & \multicolumn{1}{|c|}{0.427} & \multicolumn{1}{|c|}{\textbf{0.431}} & \multicolumn{1}{|c|}{\textbf{0.424}} & \multicolumn{1}{|c|}{0.28\%} & \multicolumn{1}{|c|}{0.69\%}\\
ETTH1\_96\_336  & \multicolumn{1}{|c|}{0.486} & \multicolumn{1}{|c|}{0.459} & \multicolumn{1}{|c|}{\textbf{0.481}} & \multicolumn{1}{|c|}{\textbf{0.454}} & \multicolumn{1}{|c|}{0.99\%} & \multicolumn{1}{|c|}{1.25\%}\\
ETTH1\_96\_720\textsuperscript{†}  & \multicolumn{1}{|c|}{0.505} & \multicolumn{1}{|c|}{0.485} & \multicolumn{1}{|c|}{\textbf{0.500}} & \multicolumn{1}{|c|}{\textbf{0.484}} & \multicolumn{1}{|c|}{0.86\%} & \multicolumn{1}{|c|}{0.20\%}\\
\midrule
ETTH2\_96\_96  & \multicolumn{1}{|c|}{0.291} & \multicolumn{1}{|c|}{0.342} & \multicolumn{1}{|c|}{\textbf{0.287}} & \multicolumn{1}{|c|}{\textbf{0.334}} & \multicolumn{1}{|c|}{1.24\%} & \multicolumn{1}{|c|}{2.60\%}\\
ETTH2\_96\_192  & \multicolumn{1}{|c|}{0.364} & \multicolumn{1}{|c|}{0.392} & \multicolumn{1}{|c|}{\textbf{0.363}} & \multicolumn{1}{|c|}{\textbf{0.389}} & \multicolumn{1}{|c|}{0.26\%} & \multicolumn{1}{|c|}{0.87\%}\\
ETTH2\_96\_336  & \multicolumn{1}{|c|}{0.418} & \multicolumn{1}{|c|}{0.426} & \multicolumn{1}{|c|}{\textbf{0.396}} & \multicolumn{1}{|c|}{\textbf{0.416}} & \multicolumn{1}{|c|}{5.76\%} & \multicolumn{1}{|c|}{2.35\%}\\
ETTH2\_96\_720\textsuperscript{†}  & \multicolumn{1}{|c|}{0.434} & \multicolumn{1}{|c|}{0.447} & \multicolumn{1}{|c|}{\textbf{0.416}} & \multicolumn{1}{|c|}{\textbf{0.440}} & \multicolumn{1}{|c|}{4.37\%} & \multicolumn{1}{|c|}{1.71\%}\\
\midrule
\textbf{AVG} & \multicolumn{1}{|c|}{0.368} & \multicolumn{1}{|c|}{0.387} & \multicolumn{1}{|c|}{\textbf{0.363}} & \multicolumn{1}{|c|}{\textbf{0.381}} & \multicolumn{1}{|c|}{1.19\%} & \multicolumn{1}{|c|}{1.52\%}\\

\bottomrule
\end{tabular}
}
\end{sc}
\end{small}
\end{center}
\caption{Regression accuracies for AdaPRL on 4 subsets of the ETT dataset using Minusformer as the backbone model. The experiments of AdaPRL were repeated three times, and the average of the corresponding metrics was reported. Sign\textsuperscript{†} indicated using a 50\% randomized sparse confidence pairwise learning due to memory bottleneck.}
\label{minus-table-ETT}
\vskip -0.1in
\end{table}

\renewcommand\arraystretch{1.2}
\begin{table}[t]
\vskip 0.15in
\begin{center}
\begin{small}
\begin{sc}
\resizebox{1.0\columnwidth}{!}{
\begin{tabular}{lcccccc}
\toprule
Model & \multicolumn{2}{|l|}{iTransformer} & \multicolumn{2}{|l|}{iTrans.+AdaPRL} & \multicolumn{2}{|l|}{Relative Improv.} \\
\toprule
Dataset & \multicolumn{1}{|c|}{MSE$\downarrow$} & \multicolumn{1}{|c|}{MAE$\downarrow$} & \multicolumn{1}{|c|}{MSE$\downarrow$} & \multicolumn{1}{|c|}{MAE$\downarrow$}& \multicolumn{1}{|c|}{MSE$\downarrow$} & \multicolumn{1}{|c|}{MAE$\downarrow$}  \\
\toprule
ECL\_96\_96  & \multicolumn{1}{|c|}{0.148} & \multicolumn{1}{|c|}{0.239} & \multicolumn{1}{|c|}{\textbf{0.147}} & \multicolumn{1}{|c|}{\textbf{0.236}} & \multicolumn{1}{|c|}{0.65\%} & \multicolumn{1}{|c|}{1.63\%}\\ 
ECL\_96\_192\textsuperscript{†}  & \multicolumn{1}{|c|}{0.167} & \multicolumn{1}{|c|}{0.258} & \multicolumn{1}{|c|}{\textbf{0.162}} & \multicolumn{1}{|c|}{\textbf{0.250}} & \multicolumn{1}{|c|}{2.85\%} & \multicolumn{1}{|c|}{3.12\%}\\
ECL\_96\_336\textsuperscript{†} & \multicolumn{1}{|c|}{0.179} & \multicolumn{1}{|c|}{0.272} & \multicolumn{1}{|c|}{\textbf{0.177}} & \multicolumn{1}{|c|}{\textbf{0.266}} & \multicolumn{1}{|c|}{1.18\%} & \multicolumn{1}{|c|}{2.31\%}\\
ECL\_96\_720\textsuperscript{†}  & \multicolumn{1}{|c|}{0.211} & \multicolumn{1}{|c|}{0.300} & \multicolumn{1}{|c|}{\textbf{0.207}} & \multicolumn{1}{|c|}{\textbf{0.293}} & \multicolumn{1}{|c|}{2.05\%} & \multicolumn{1}{|c|}{2.24\%}\\
\midrule
\textbf{AVG} & \multicolumn{1}{|c|}{0.176} & \multicolumn{1}{|c|}{0.267} & \multicolumn{1}{|c|}{\textbf{0.173}} & \multicolumn{1}{|c|}{\textbf{0.261}} & \multicolumn{1}{|c|}{1.73\%} & \multicolumn{1}{|c|}{2.33\%}\\
\bottomrule
\end{tabular}
}

\end{sc}
\end{small}
\caption{Regression accuracies for AdaPRL on the Electricity dataset using iTransformer as the backbone model. The experiments of AdaPRL were repeated three times, and the average of the corresponding metrics was reported. Sign\textsuperscript{†} indicated using a randomized sparse confidence pairwise learning due to memory bottleneck.}
\label{itransformer-table-ECL}
\end{center}
\vskip -0.1in
\end{table}

\renewcommand\arraystretch{1.2}
\begin{table}[t]

\vskip 0.15in
\begin{center}
\begin{small}
\begin{sc}
\resizebox{1.0\columnwidth}{!}{
\begin{tabular}{lcccccc}
\toprule
Model & \multicolumn{2}{|l|}{Minusformer} & \multicolumn{2}{|l|}{Minus.+AdaPRL} & \multicolumn{2}{|l|}{Relative Improv.} \\
\toprule
Dataset & \multicolumn{1}{|c|}{MSE$\downarrow$} & \multicolumn{1}{|c|}{MAE$\downarrow$} & \multicolumn{1}{|c|}{MSE$\downarrow$} & \multicolumn{1}{|c|}{MAE$\downarrow$}& \multicolumn{1}{|c|}{MSE$\downarrow$} & \multicolumn{1}{|c|}{MAE$\downarrow$}  \\
\toprule
ECL\_96\_96  & \multicolumn{1}{|c|}{0.143} & \multicolumn{1}{|c|}{0.235} & \multicolumn{1}{|c|}{\textbf{0.142}} & \multicolumn{1}{|c|}{\textbf{0.230}} & \multicolumn{1}{|c|}{1.00\%} & \multicolumn{1}{|c|}{2.05\%}\\ 
ECL\_96\_192\textsuperscript{†}   & \multicolumn{1}{|c|}{0.160} & \multicolumn{1}{|c|}{0.251} & \multicolumn{1}{|c|}{\textbf{0.160}} & \multicolumn{1}{|c|}{\textbf{0.248}} & \multicolumn{1}{|c|}{0.07\%} & \multicolumn{1}{|c|}{1.48\%}\\
ECL\_96\_336\textsuperscript{†}   & \multicolumn{1}{|c|}{0.174} & \multicolumn{1}{|c|}{0.267} & \multicolumn{1}{|c|}{\textbf{0.174}} & \multicolumn{1}{|c|}{\textbf{0.263}} & \multicolumn{1}{|c|}{0.15\%} & \multicolumn{1}{|c|}{1.52\%}\\
ECL\_96\_720\textsuperscript{†}  & \multicolumn{1}{|c|}{0.205} & \multicolumn{1}{|c|}{0.204} & \multicolumn{1}{|c|}{\textbf{0.202}} & \multicolumn{1}{|c|}{\textbf{0.288}} & \multicolumn{1}{|c|}{1.82\%} & \multicolumn{1}{|c|}{2.13\%}\\
\midrule
\textbf{AVG} & \multicolumn{1}{|c|}{0.171} & \multicolumn{1}{|c|}{0.262} & \multicolumn{1}{|c|}{\textbf{0.169}} & \multicolumn{1}{|c|}{\textbf{0.257}} & \multicolumn{1}{|c|}{0.74\%} & \multicolumn{1}{|c|}{1.80\%}\\

\bottomrule
\end{tabular}
}
\end{sc}
\end{small}
\end{center}
\caption{Regression accuracies for AdaPRL on the ECL dataset using Minusformer as the backbone model. The experiments of AdaPRL were repeated three times, and the average of the corresponding metrics was reported. Sign\textsuperscript{†} indicated using a randomized sparse confidence pairwise learning due to memory bottleneck.}
\label{minus-table-ECL}
\vskip -0.1in
\end{table}

\subsubsection{Overall Performance}

As shown in Table \ref{itransformer-table-ETT}, Table \ref{minus-table-ETT}, Table \ref{itransformer-table-ECL} and Table \ref{minus-table-ECL}, comparing with using a single MSE loss, our proposed method AdaPRL significantly improved the precision across all prediction lengths for every subset of the ETT dataset (ETTm1, ETTm2, ETTh1, ETTh2) as well as the ECL dataset. The experimental results also demonstrated that AdaPRL worked well with different model backbones, including two state-of-the-art time series forecasting models, iTransformer and Minusformer, and boosted their performance to even higher levels. AdaPRL improved the MSE and MAE by 1.65\% and 1.86\% with iTransformer, and 1.19\% and 1.52\% with Minusformer on the entire ETT dataset. On the ECL dataset, AdaPRL improved the MSE and MAE by 1.73\% and 2.33\% with iTransformer, and 0.74\% and 1.80\% with Minusformer, respectively. The visualizations of the forecasting results and uncertainty prediction for these datasets and models are provided in section \ref{vis-explain-section}.

\subsection{Age Prediction}
\subsubsection{Datasets}
\textbf{The AgeDB\footnote{https://ibug.doc.ic.ac.uk/resources/agedb/}} \cite{moschoglou2017agedb} is a dataset released by Sberbank AI Lab, focused on advancing research in cross-age facial recognition. It consists of 16,488 facial images, each annotated with gender and age. We use the RetinaFace model for face detection and cropping to ensure each image fully captures the face. The images are resized to 256x256 pixels, and the pixel values are normalized. The dataset is split into training, testing, and validation sets in a 6:2:2 ratio.

\textbf{The CLAP2016\footnote{https://chalearnlap.cvc.uab.cat/dataset/26/description/}} \cite{escalera2016chalearn} dataset is a large-scale facial image database designed for the estimation of both real and apparent age. It contains 7,591 facial images, each annotated with the corresponding real and apparent age. According to the official data format, these images are divided into 4,113 training images, 1,500 validation images, and 1,978 test images. Similarly, we employ the RetinaFace model for facial image detection and cropping, and apply the same preprocessing techniques as those used in AgeDB for subsequent image processing.

\subsubsection{Experiment Setup}
Following the comparison strategy proposed by the Facial Age Estimation Benchmark \cite{paplham2024call}, we adapted ResNet50 \cite{he2016deep} pre-trained in ImageNet \cite{deng2009imagenet} as the backbone and adopted MAE as the loss function. Each training session was run for a total of 100 epochs. We repeated each experiment five times and reported the MAE and Kendall’s tau. A coarse-grained grid search was performed on the hyperparameters, including the pairwise ranking loss type and the weight factor $\alpha$ from the set \{0.5,1.0,2.0,4.0,8.0,16.0,32.0,64.0\}.

\subsubsection{Overall Performance}
The detailed experimental results are delineated in Table \ref{age-est}. We highlighted the highest-performing datasets in bold and the second-best in underlined text. Unlike other regression tasks that can concurrently evaluate MSE and MAE metrics, age estimation typically prioritizes the MAE metric. We augmented L1, Huber, and L2 loss functions with both original pairwise learning  (L2+PRL) and AdaPRL to assess performance across two datasets. The experimental findings yielded the following insights: within the AgeDB and CLAP2016 datasets, AdaPRL consistently attained the most exemplary performance irrespective of the baseline regression loss employed. Specifically, on the AgeDB dataset, the MAE metric was reduced by 1.13\%, 0.36\%, and 0.71\% when combined with the L1, Huber, and L2 regressions, respectively; on the CLAP2016 dataset, the reductions were 7.47\%, 6.13\%, and 4.13\%, respectively. Additionally, while achieving more accurate age estimations, both PRL and AdaPRL also gained improvements in the Kendall's \(\tau\) across both datasets. This demonstrated that our approach excelled in preserving the ordinal relationships of actual ages, thereby more precisely reflecting the relative age dynamics among individuals.

\begin{table}[h]
\begin{center}
\renewcommand{\arraystretch}{1.2} 
\begin{small}
\begin{sc}\resizebox{0.98\columnwidth}{!}{%
\begin{tabular}{l|cc|cc}
\toprule
Dataset & \multicolumn{2}{c|}{AgeDB} & \multicolumn{2}{c}{CLAP2016} \\ 
\cmidrule(lr){1-5}
Model & MAE$\downarrow$ & Kendall's \(\tau\)$\uparrow$ & MAE$\downarrow$ & Kendall's \(\tau\)$\uparrow$ \\ 
\midrule
L1 & 6.372 & 0.697 & 4.792 & 0.753 \\ 
L1+PRL & \underline{6.323} & \underline{0.701} & \underline{4.507} & \underline{0.763} \\ 
AdaPRL(L1) & \textbf{6.300} & \textbf{0.702} & \textbf{4.434} & \textbf{0.765} \\ 
\midrule
\emph{Rel.Imp} & 1.13\% & 0.72\% & 7.47\% & 1.59\% \\
\midrule
Huber & 6.333 & 0.700 &4.680&0.753  \\ 
Huber+PRL & \underline{6.313} & \underline{0.701} & \underline{4.444}&\underline{0.770}  \\ 
AdaPRL(Huber) & \textbf{6.310} & \textbf{0.702} & \textbf{4.393} &\textbf{0.778}  \\ 
\midrule
\emph{Rel.Imp} & 0.36\% & 0.29\% & 6.13\%&3.32\% \\
\midrule
L2 & 6.372  & 0.697  & 4.726&	0.753  \\ 
L2+PRL & \underline{6.353} & \underline{0.698}  & \underline{4.545}&	\underline{0.757}  \\ 
AdaPRL(L2) & \textbf{6.327}&\textbf{0.699}  & \textbf{4.531}&\textbf{0.759}  \\ 
\midrule
\emph{Rel.Imp} & 0.71\%&0.29\% & 4.13\%& 0.80\% \\
\bottomrule
\end{tabular}%
}
\end{sc}
\end{small}
\caption{Comparison of various methods' performance on the AgeDB and CLAP2016 datasets. \emph{Rel.Imp} denotes the relative improvement over the corresponding metrics.}
\label{age-est}
\end{center}
\end{table}

\subsection{Natural Language Understanding}
\subsubsection{Datasets}
\textbf{Jigsaw\footnote{https://www.kaggle.com/datasets/nkitgupta/jigsaw-regression-based-data/data}} \cite{jigsaw-toxic-severity-rating} is a dataset of natural language understanding released for the purpose of studying the level of all kinds of negative toxic comments (i.e. comments that are rude, disrespectful or otherwise likely to cause someone to leave a discussion). The target is to predict the level of toxicity of each annotated user comment sample. We formalize the task as a regression problem and the data is split randomly into 80\% for training, 10\% for validation, and 10\% for testing.
\subsubsection{Experiment Setup}
In our experiments, we employed a pre-trained RoBERTa \cite{liu_roberta_2019} base model as the baseline backbone. To better facilitate training in larger batch sizes and improve accuracy, we used the LAMB \cite{you_large_2020} optimizer with a batch size of 320 and a learning rate of 8e-4 using a warm-up cosine annealing scheduler. We implemented an early stopping strategy based on a patience of 2 in the validation set. A grid search was conducted on the hyperparameters including ranking loss type, \(\alpha\) between \{0.001,0.002, 0.005, 0.01,0.02,0.05,0.1,0.2\} and \(\theta\) between \{0,0.3,0.9\}. The experiments were carried out over 10 runs to reduce randomness, and the average performance on the metrics of the test set was reported.

\subsubsection{Overall Performance}
The detailed experimental results are presented in Table \ref{jigsaw-table}. Similarly, we emphasized the top-performing datasets in bold and denoted the second-best performing datasets with underlines. The findings demonstrated that the proposed AdaPRL(L2) loss function excelled, achieved the lowest MSE of 0.3561 and MAE of 0.2922, and represented improvements of 3.39\% and 2.92\% over the standard L2 loss respectively. Although the original pairwise learning (L2+PRL) did not match AdaPRL's performance, it still offered a significant enhancement over the L2 loss in both MSE and MAE. Additionally, we can also observe that both AdaPRL and original pairwise learning (L2+PRL) attained a higher Kendall's \(\tau\) (0.6858) compared to L2, further illustrating that our approach more effectively maintains the relative ranking among samples when predicting the toxicity levels of user comments, which is crucial for accurately assessing the severity of reviews.

\begin{table}[t]
\vskip 0.15in
\begin{center}
\begin{small}
\begin{sc}\resizebox{0.9\columnwidth}{!}{
\begin{tabular}{lcccc}
\toprule
Dataset & loss type & MSE$\downarrow$ & MAE$\downarrow$ & Kendall‘s \(\tau\)$\uparrow$ \\
\midrule

\multirow{4}{*}[-1.5ex]{\textbf{JigSaw}}   

  & L2 & 0.3686& 0.3010 & 0.6843 \\ 
  & L2+PRL & \underline{0.3569}& \underline{0.2988} & \underline{0.6858}\\
  & AdaPRL(L2) & \textbf{0.3561}& \textbf{0.2922} & \textbf{0.6858}\\
  \cmidrule(lr){2-5}
  & \emph{Rel.Imp} & 3.39\% & 2.92\% & 0.22\%\\
\bottomrule
\end{tabular}
}
\end{sc}
\end{small}
\caption{Regression accuracies for Pairwise Regression Learning and AdaPRL on Jigsaw. \emph{Rel.Imp} indicates relative improvement over corresponding metrics. The experiments were repeated 10 times, and the average results were reported.}
\label{jigsaw-table}
\end{center}
\vskip -0.1in
\end{table}

\subsection{Financial Datasets}
\subsubsection{Datasets}
\textbf{Janestreet\footnote{https://www.kaggle.com/competitions/jane-street-real-time-market-data-forecasting/data}} \cite{jane-street-real-time-market-data-forecasting} data comprise a set of time series with 79 anonymized features, 1 sample weight column, and 9 responders, representing real market data. The goal is to forecast one of the responders for the future. We formalize the task as a univariate regression problem, skip the first 500 days of data to avoid missing data, and randomly split 50\% / 50\% of the data as validation and test set for the last 100 days. Following the data description and the official evaluation protocol, we use weighted \(R^2\) as evaluation metrics.
\subsubsection{Experiment Setup}
In our experiments, we employed a multi-head attention model \cite{vaswani_attention_2023} as encoder followed by a multilayer perceptron as the baseline backbone. The input sequence length, stride, and prediction length were set to 64, 16, and 1 respectively, and the model was trained on respond\_6 as a single target. All experiments used AdamW \cite{loshchilov2017decoupled} as optimizer and the weighted MSE as \(L_{reg}\). We implemented an early stopping strategy based on a patience of 3 in the validation set. A grid search was conducted on the hyperparameters including ranking loss type, \(\alpha\) between \{0.005, 0.01,0.02,0.04,0.08,0.16,0.32\}. The experiments were conducted over 3 runs and we reported the average weighted \(R^2\) metrics on the test set.

\subsubsection{Overall Performance}
The comprehensive experimental results are delineated in Table \ref{js-table}. The premier method AdaPRL attained the highest weighted \( R^2 \) score of 0.00617, thereby eclipsing alternative approaches. In comparison to the baseline loss function weighted MSE, which yielded a weighted \( R^2 \) of 0.00540, AdaPRL demonstrated a relative enhancement of 14.81\%. In addition, the pairwise learning technique weighted MSE+PRL achieved a marginally superior performance of 0.00575, yet it did not match the precision of AdaPRL. These findings substantiate that AdaPRL effectively increases regression accuracy, surpassing both the conventional weighted MSE and the standard pairwise learning-based approaches.

\begin{table}[t]
\vskip 0.15in
\begin{center}
\begin{small}
\begin{sc}\resizebox{0.825\columnwidth}{!}{
\begin{tabular}{lcc}
\toprule
Dataset & loss type & weighted $R^2$$\uparrow$  \\
\midrule
\multirow{3}{*}[-1.5ex]{\textbf{Janestreet}}  
                        & Weighted MSE & 0.00540 \\ 
                        & Weighted MSE+PRL & \underline{0.00575} \\
                        & AdaPRL & \textbf{0.00617}\\
  \cmidrule(lr){2-3}
  & \emph{Rel.Imp} & 14.81\%  \\
\bottomrule
\end{tabular}
}
\end{sc}
\end{small}
\caption{Regression accuracies for Pairwise Regression Learning (PRL) and AdaPRL on Janestreet. \emph{Rel.Imp} indicates relative improvement over corresponding metrics. The experiments were repeated three times, and the average results were reported.}
\label{js-table}
\end{center}
\vskip -0.1in
\end{table}

\subsection{Industrial Datasets}
\subsubsection{Datasets and Experiment Setup}
We evaluated our method on a large-scale proprietary industrial regression dataset from a major music streaming app. It contained millions of records with over 300 sparse features and the task was to predict the expected income of each user in the next day for commercial application. We split the data according to the actual timeline: the first 14 days were used as the training set, while the last day was evenly divided to create validation and test sets. In our experiments, all features were processed as sparse features, and we used a DNN model with a feature embedding layer as the baseline to compare different loss functions. The model was trained for a single epoch using the Adam optimizer with a learning rate of 1e-4. A coarse-grained grid search was performed on the hyperparameter, including the type of ranking loss and \(\alpha\) between \{0.01,0.02,0.05,0.1,0.2\}. 

\subsubsection{Overall Performance}
As shown in Table \ref{qm-table}, AdaPRL achieved the highest accuracy among various loss functions on our industrial dataset, demonstrating the effectiveness of the proposed training techniques. Native pairwise regression learning improved MSE and MAE by 0.51\% and 1.60\%, respectively. In contrast, the AdaPRL method, leveraging pairwise learning and uncertainty estimation, achieved total gains of 0.96\% and 1.93\% in MSE and MAE. In particular, in large-scale music streaming applications with tens of millions of users, even a modest improvement in accuracy of 0.1\% is empirically significant and can positively impact business outcomes. The proposed method showcases substantial superiority in capturing and utilizing the relative relationships within user data. Furthermore, Kendall's \(\tau\) score for the proposed approach is the highest among the evaluated loss functions, reflecting enhanced ranking capabilities, which are crucial for identifying high-value users in real-world applications.

\begin{table}[t]
\vskip 0.15in
\begin{center}
\begin{small}
\begin{sc}\resizebox{0.72\columnwidth}{!}{
\begin{tabular}{lccc}
\toprule
Dataset & loss type & MSE$\downarrow$ & MAE$\downarrow$  \\
\midrule
\multirow{3}{*}[-1.5ex]{\textbf{Industrial}}  
                        & L2 & 0.5301 & 0.2941 \\ 
                        & L2+PRL & \underline{0.5274} & \underline{0.2894} \\
                        & AdaPRL & \textbf{0.5250} & \textbf{0.2838}\\
  \cmidrule(lr){2-4}
  & \emph{Rel.Imp} & 0.96\% & 1.93\%  \\
\bottomrule
\end{tabular}
}
\end{sc}
\end{small}
\caption{Regression accuracies for Pairwise Regression Learning (PRL) and AdaPRL on large-scale proprietary industrial regression dataset from a major music streaming app. \emph{Rel.Imp} indicates relative improvement over corresponding metrics. The experiments were repeated three times and the average results were reported.}
\label{qm-table}
\end{center}
\vskip -0.1in
\end{table}

\section{Further Study and Analysis}

\subsection{Sensitivity of Hyperparameter Alpha}
\subsubsection{Experiment Setup}
To investigate the impact and sensitivity of the hyperparameter in AdaPRL, we conducted experiments on the hyperparameter alpha in this section. We chose the MovieLens-1M, KuaiRec, and ETT datasets for analysis. We employed an RMSE type AdaPRL as the auxiliary loss for the MovieLens-1M dataset and MAE type AdaPRL for all the other datasets. Note that \(\theta\) was set to 0 for all experiments. All other experimental settings remained consistent with those described in \ref{app_sec_a}.

\begin{figure}[ht]
\vskip 0.2in
\begin{center}
    \begin{minipage}{0.9\columnwidth}
        \centering
        \includegraphics[width=\linewidth]{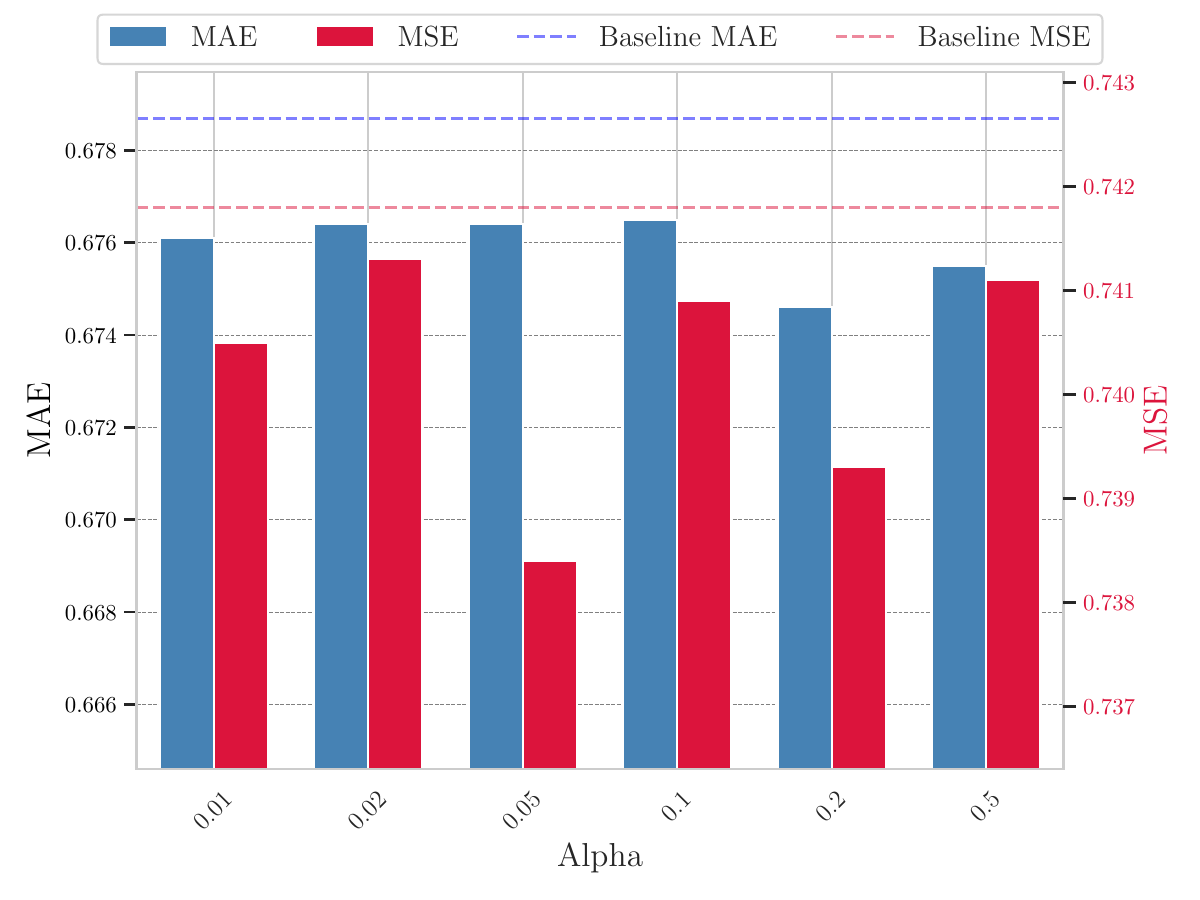}
    \end{minipage}
    \hfill
    \begin{minipage}{0.9\columnwidth}
        \centering
        \includegraphics[width=\linewidth]{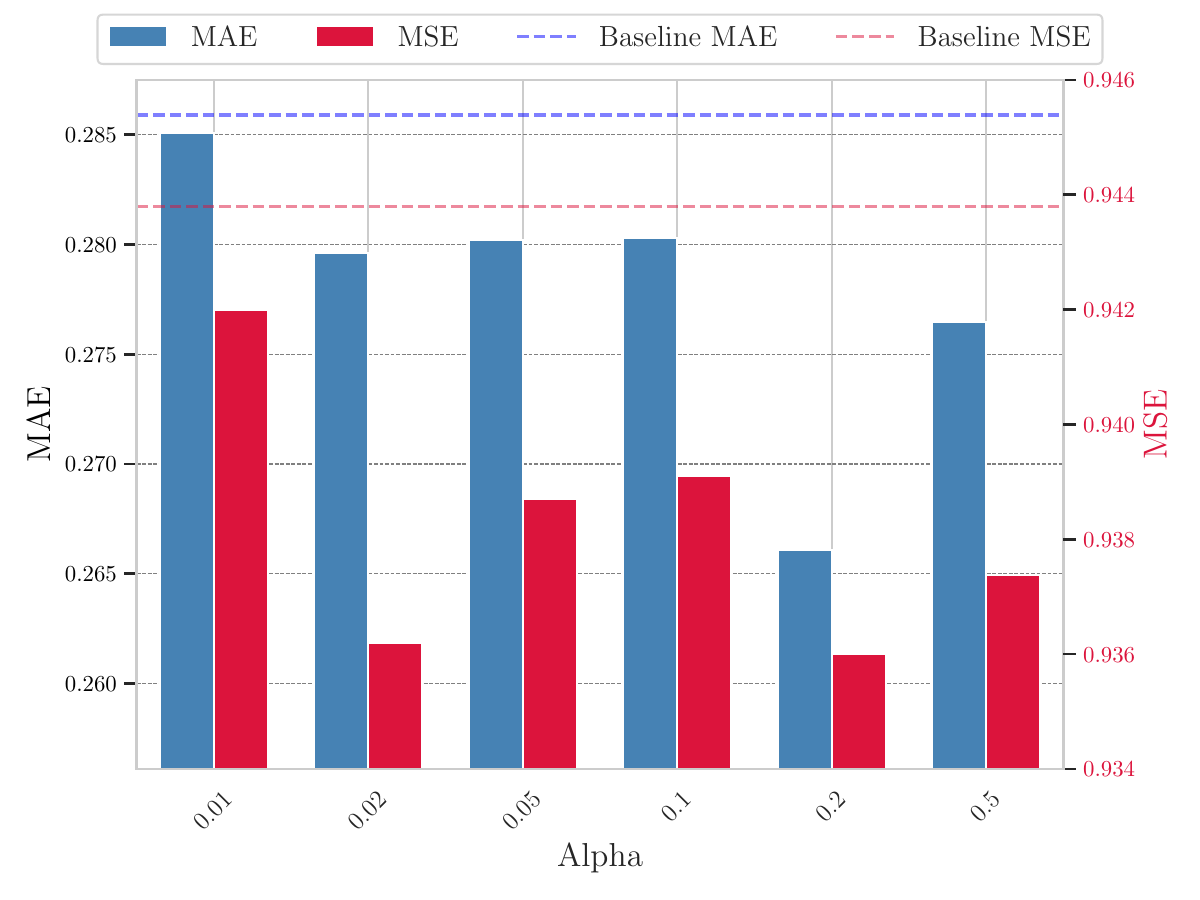}
    \end{minipage}
    \caption{Influence of hyperparameter alpha in Movielens-1M and KuaiRec datasets. The upper subplot illustrated the evaluation metrics in Movielens-1M, while the lower one presented the experimental results in KuaiRec.}

    \label{rec-alpha-abalation}
\end{center}
\vskip -0.2in
\end{figure}

\begin{figure}[ht]
\vskip 0.2in
\begin{center}
    \begin{minipage}{0.9\columnwidth}
        \centering
        \includegraphics[width=\linewidth]{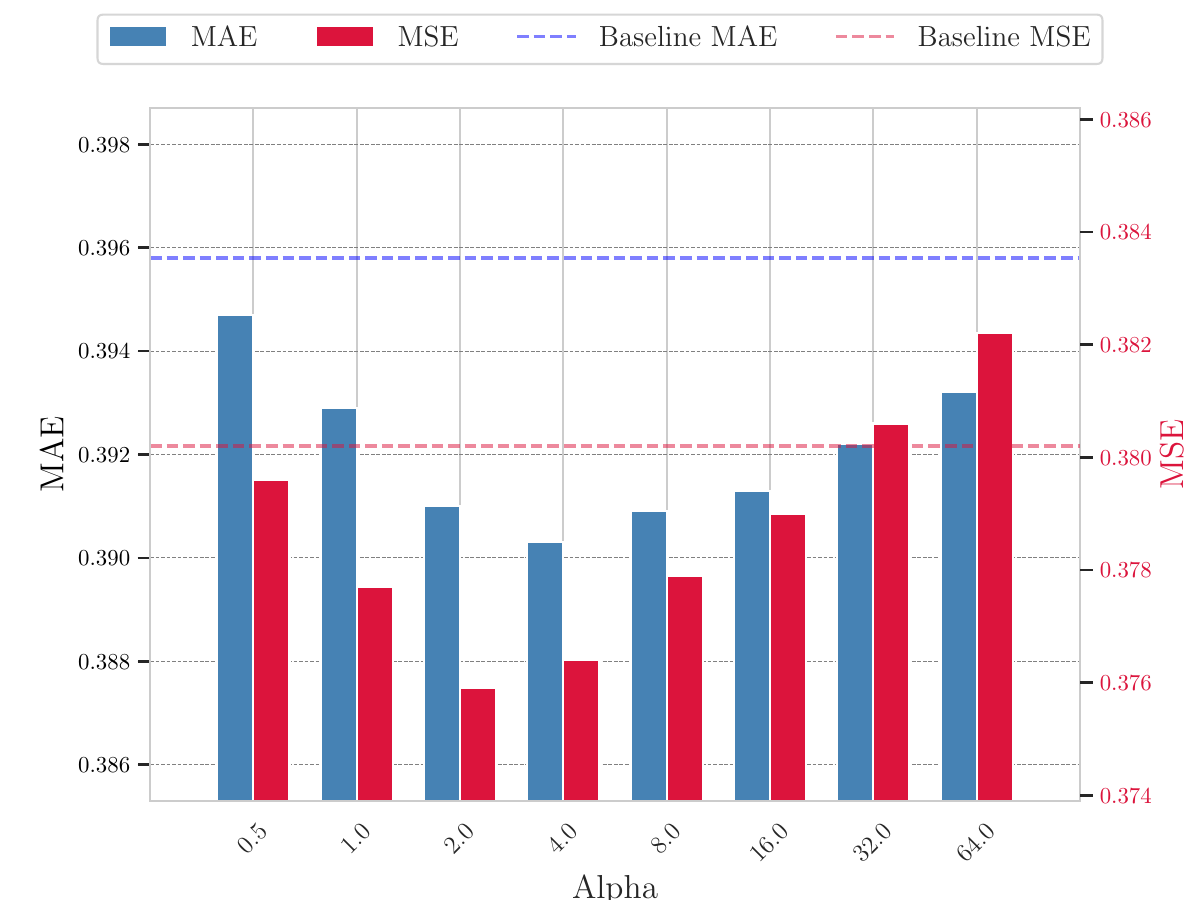}
    \end{minipage}
    \hfill
    \begin{minipage}{0.89\columnwidth}
        \centering
        \includegraphics[width=\linewidth]{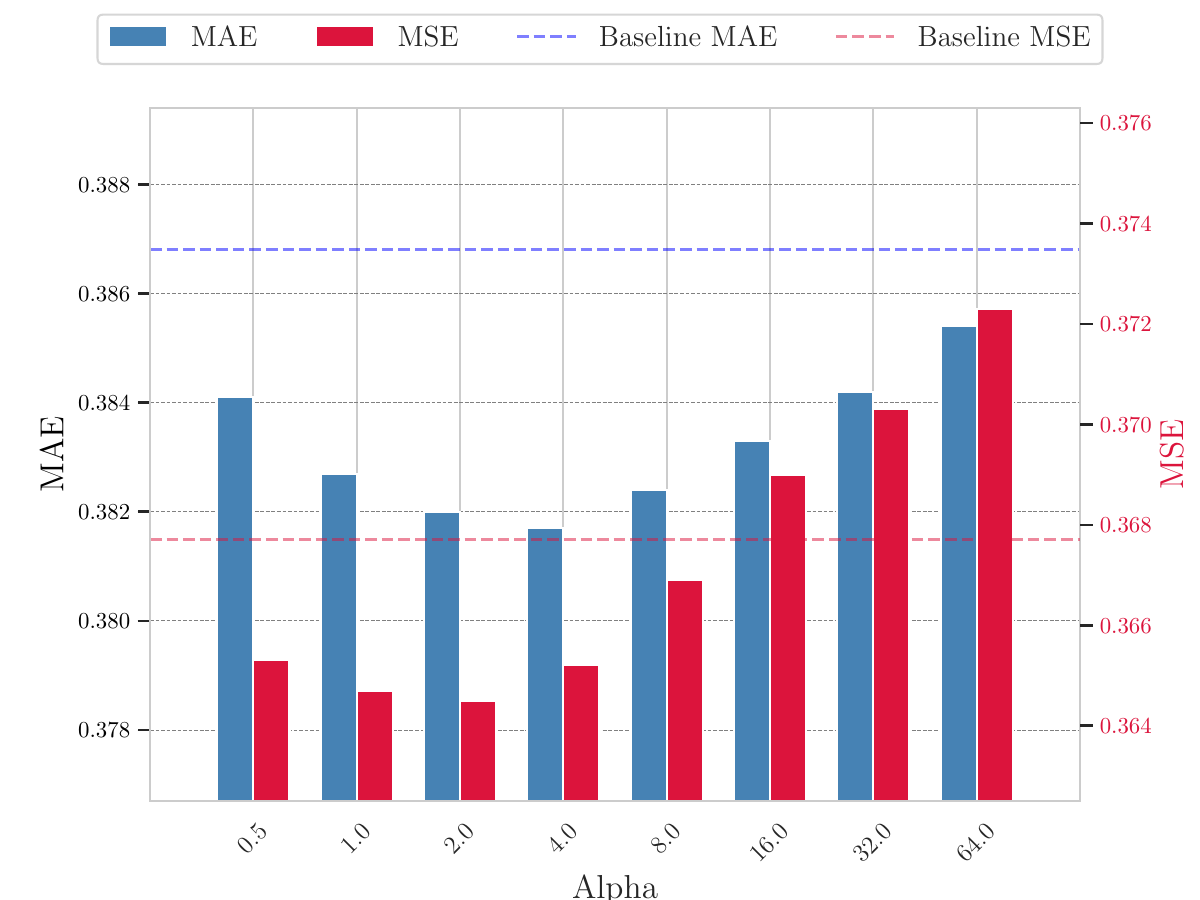}
    \end{minipage}
    \caption{Influence of hyperparameter alpha in the ETT dataset. The upper subplot illustrated the evaluation metrics using iTransformer as the backbone network, while the lower one presented the experimental results using Minusformer as the backbone network.}
    
    \label{ett-alpha-abalation}
\end{center}
\vskip -0.2in
\end{figure}

\subsubsection{Results}
The results are shown in Figures \ref{rec-alpha-abalation} and \ref{ett-alpha-abalation}. As illustrated in the figures, AdaPRL consistently achieved higher precision than the regression baseline on the MovieLens-1M, KuaiRec, and ETT datasets in most choices of alpha. Moreover, as shown in our earlier experiment sections, a coarse-grained search on alpha yielded a significant improvement with AdaPRL on multiple regression datasets. These results suggest that our proposed AdaPRL approach is robust and not highly sensitive to the choice of the alpha hyperparameter.

\subsection{Effect of Sparsity in AdaPRL}
\subsubsection{Experiment Setup}
As we analyzed before in \ref{SCPRL}, if randomness is induced for \(L_{SCPRL}\) when generating the sparse area \(S\), this method could be viewed as a variant of dropout applied to confidence pairwise learning. Therefore, we found it essential to perform a quantitative analysis of the potential effects of dropout on the algorithm and empirically investigated how the underlying mechanisms of random dropout influenced the model. The evaluation was conducted on a subset of the ETT dataset, specifically focusing on prediction tasks with a forecasting horizon of 720 time steps in this experiment to analyze since the problem of the bottleneck of memory and computation was mainly raised in the scenario of multivariate time series forecasting while incorporating AdaPRL. We used iTransformer as the baseline model and conducted a grid search of sparsity based on random masking techniques; hyperparameter alpha was optimized via a coarse-grained grid search, and the best result was reported. All other settings remained the same as were described above in \ref{ts-exp-setup}. 

\begin{figure}[ht]
\vskip 0.2in
\begin{center}
\begin{minipage}{1.0\columnwidth}
\centerline{\includegraphics[width=\columnwidth]{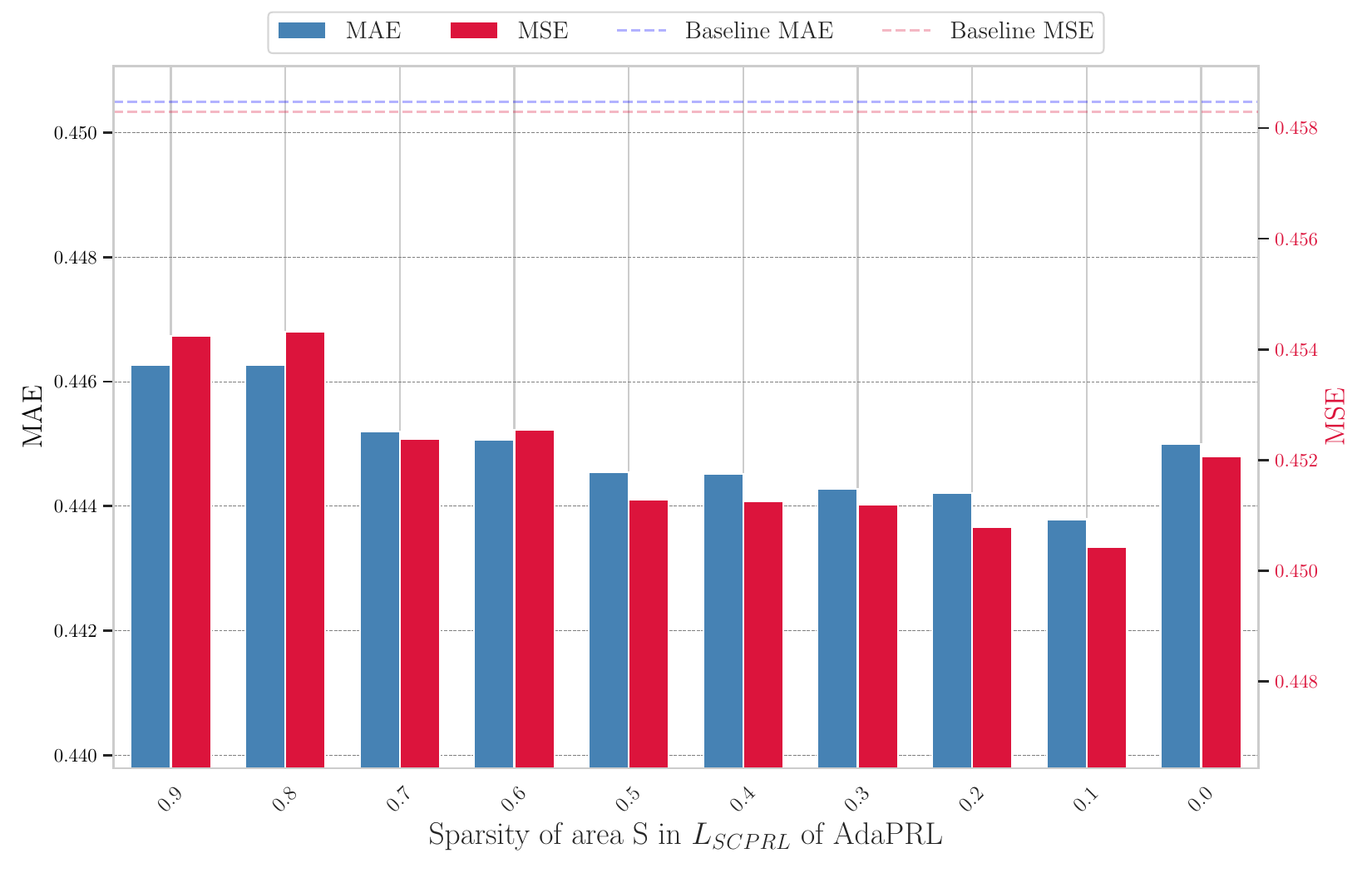}}
\caption{The average performance of our AdaPRL with different levels of sparsity in the ETT dataset subset with a forecasting horizon of 720 time steps.}

\label{sparsity-ett}
\end{minipage}
\end{center}
\vskip -0.2in
\end{figure}

\subsubsection{Results}
The result is shown in Figure \ref{sparsity-ett}. It is obvious that as the sparsity of the area gradually decreases from 0.9 to 0.1, where computation gradually increases, the algorithm's evaluation loss MAE and MSE exhibited an overall downward trend. This indicates that reducing sparsity improves the model's predictive accuracy and robustness, likely by preserving essential information and mitigating the loss of critical features. However, when the sparsity is set to 1.0, corresponding to using a non-sparse algorithm, both MAE and MSE are worse than those observed at most other sparsity levels. This suggests that the introduction of a certain (probably relatively lower) level of sparsity, corresponding to a small drop rate in dropout, plays a crucial role in preventing overfitting and reducing computational complexity, thereby improving generalization performance. Conversely, the absence of sparsity, although it computes the relative difference of more pairs of data and requires more memory consumption, still leads to suboptimal performance compared with a sparse version of AdaPRL. The complete result of the sparsity test of the ETT dataset is listed in Table \ref{trans-table-sparsity}.

\renewcommand\arraystretch{1.2}
\begin{table}[t]
\vskip 0.15in
\begin{center}
\begin{small}
\begin{sc}
\resizebox{1.0\columnwidth}{!}{
\begin{tabular}{lccclccc}
\toprule
\multicolumn{1}{c}{Dataset} & \multicolumn{1}{|c|}{Sparsity} & \multicolumn{1}{|c|}{MSE$\downarrow$} & \multicolumn{1}{|c|}{MAE$\downarrow$} & \multicolumn{1}{|c|}{Dataset} & \multicolumn{1}{|c|}{Sparsity} & \multicolumn{1}{|c|}{MSE$\downarrow$} & \multicolumn{1}{|c|}{MAE$\downarrow$}  \\
\toprule

\multirow{10}{*}[-1.5ex]{\textbf{ETTM1\_96\_720}}  
  & \multicolumn{1}{|c|}{0.0} & \multicolumn{1}{|c|}{0.4815} & \multicolumn{1}{|c|}{0.4485}  & \multirow{10}{*}[-1.5ex]{\textbf{ETTM2\_96\_720}} & \multicolumn{1}{|c|}{0.0} & \multicolumn{1}{|c|}{0.4047} & \multicolumn{1}{|c|}{0.3989} \\
  & \multicolumn{1}{|c|}{0.1} & \multicolumn{1}{|c|}{\underline{0.4812}} & \multicolumn{1}{|c|}{0.4494}  & & \multicolumn{1}{|c|}{0.1} & \multicolumn{1}{|c|}{0.4049} & \multicolumn{1}{|c|}{0.3981} \\
  & \multicolumn{1}{|c|}{0.2} & \multicolumn{1}{|c|}{0.4814} & \multicolumn{1}{|c|}{\underline{0.4487}}  & & \multicolumn{1}{|c|}{0.2} & \multicolumn{1}{|c|}{0.4044} & \multicolumn{1}{|c|}{0.3986} \\
  & \multicolumn{1}{|c|}{0.3} & \multicolumn{1}{|c|}{0.4815} & \multicolumn{1}{|c|}{0.4489}  & & \multicolumn{1}{|c|}{0.3} & \multicolumn{1}{|c|}{0.4056} & \multicolumn{1}{|c|}{0.3989} \\
  & \multicolumn{1}{|c|}{0.4} & \multicolumn{1}{|c|}{\textbf{0.4809}} & \multicolumn{1}{|c|}{\textbf{0.4485}}  & & \multicolumn{1}{|c|}{0.4} & \multicolumn{1}{|c|}{\underline{0.4030}} & \multicolumn{1}{|c|}{0.3975} \\
  & \multicolumn{1}{|c|}{0.5} & \multicolumn{1}{|c|}{0.4812} & \multicolumn{1}{|c|}{0.4490}  & & \multicolumn{1}{|c|}{0.5} & \multicolumn{1}{|c|}{0.4038} & \multicolumn{1}{|c|}{0.3981} \\
  & \multicolumn{1}{|c|}{0.6} & \multicolumn{1}{|c|}{0.4812} & \multicolumn{1}{|c|}{0.4491}  & & \multicolumn{1}{|c|}{0.6} & \multicolumn{1}{|c|}{0.4039} & \multicolumn{1}{|c|}{\underline{0.3972}} \\
  & \multicolumn{1}{|c|}{0.7} & \multicolumn{1}{|c|}{0.4815} & \multicolumn{1}{|c|}{0.4493}  & & \multicolumn{1}{|c|}{0.7} & \multicolumn{1}{|c|}{\textbf{0.4014}} & \multicolumn{1}{|c|}{\textbf{0.3964}} \\
  & \multicolumn{1}{|c|}{0.8} & \multicolumn{1}{|c|}{0.4813} & \multicolumn{1}{|c|}{0.4495}  & & \multicolumn{1}{|c|}{0.8} & \multicolumn{1}{|c|}{0.4070} & \multicolumn{1}{|c|}{0.3997} \\
  & \multicolumn{1}{|c|}{0.9} & \multicolumn{1}{|c|}{0.4813} & \multicolumn{1}{|c|}{0.4492}  & & \multicolumn{1}{|c|}{0.9} & \multicolumn{1}{|c|}{0.4061} & \multicolumn{1}{|c|}{0.3996} \\
\midrule
\multirow{10}{*}[-1.5ex]{\textbf{ETTH1\_96\_720}}  
  & \multicolumn{1}{|c|}{0.0} & \multicolumn{1}{|c|}{0.5010} & \multicolumn{1}{|c|}{0.4890}  & \multirow{10}{*}[-1.5ex]{\textbf{ETTH2\_96\_720}} & \multicolumn{1}{|c|}{0.0} & \multicolumn{1}{|c|}{0.4211} & \multicolumn{1}{|c|}{0.4436} \\
  & \multicolumn{1}{|c|}{0.1} & \multicolumn{1}{|c|}{\textbf{0.4955}} & \multicolumn{1}{|c|}{\textbf{0.4855}}  & & \multicolumn{1}{|c|}{0.1} & \multicolumn{1}{|c|}{0.4201} & \multicolumn{1}{|c|}{0.4422} \\
  & \multicolumn{1}{|c|}{0.2} & \multicolumn{1}{|c|}{\underline{0.4966}} & \multicolumn{1}{|c|}{\underline{0.4862}}  & & \multicolumn{1}{|c|}{0.2} & \multicolumn{1}{|c|}{0.4207} & \multicolumn{1}{|c|}{0.4434} \\
  & \multicolumn{1}{|c|}{0.3} & \multicolumn{1}{|c|}{0.4974} & \multicolumn{1}{|c|}{0.4869}  & & \multicolumn{1}{|c|}{0.3} & \multicolumn{1}{|c|}{0.4203} & \multicolumn{1}{|c|}{0.4424} \\
  & \multicolumn{1}{|c|}{0.4} & \multicolumn{1}{|c|}{0.5005} & \multicolumn{1}{|c|}{0.4887}  & & \multicolumn{1}{|c|}{0.4} & \multicolumn{1}{|c|}{0.4206} & \multicolumn{1}{|c|}{0.4433} \\
  & \multicolumn{1}{|c|}{0.5} & \multicolumn{1}{|c|}{0.5008} & \multicolumn{1}{|c|}{0.4891}  & & \multicolumn{1}{|c|}{0.5} & \multicolumn{1}{|c|}{0.4193} & \multicolumn{1}{|c|}{0.4419} \\
  & \multicolumn{1}{|c|}{0.6} & \multicolumn{1}{|c|}{0.5050} & \multicolumn{1}{|c|}{0.4914}  & & \multicolumn{1}{|c|}{0.6} & \multicolumn{1}{|c|}{0.4201} & \multicolumn{1}{|c|}{0.4425} \\
  & \multicolumn{1}{|c|}{0.7} & \multicolumn{1}{|c|}{0.5074} & \multicolumn{1}{|c|}{0.4930}  & & \multicolumn{1}{|c|}{0.7} & \multicolumn{1}{|c|}{\underline{0.4192}} & \multicolumn{1}{|c|}{0.4421} \\
  & \multicolumn{1}{|c|}{0.8} & \multicolumn{1}{|c|}{0.5092} & \multicolumn{1}{|c|}{0.4940}  & & \multicolumn{1}{|c|}{0.8} & \multicolumn{1}{|c|}{0.4198} & \multicolumn{1}{|c|}{\underline{0.4419}} \\
  & \multicolumn{1}{|c|}{0.9} & \multicolumn{1}{|c|}{0.5110} & \multicolumn{1}{|c|}{0.4946}  & & \multicolumn{1}{|c|}{0.9} & \multicolumn{1}{|c|}{\textbf{0.4185}} & \multicolumn{1}{|c|}{\textbf{0.4417}} \\
  
\bottomrule
\end{tabular}
}
\end{sc}
\end{small}
\end{center}
\caption{Regression accuracies for AdaPRL on 4 subsets of ETT dataset with prediction length of 720 using iTransformer as backbone model. The experiments of AdaPRL were repeated three times and the average of corresponding metrics were reported.}
\label{trans-table-sparsity}
\vskip -0.1in
\end{table}

\subsection{Generalization and Integration with Existing Methods}
\subsubsection{Experiment Setup}
In this section, we evaluated the generalizability of AdaPRL and its compatibility with previously proposed methods. As shown in Table \ref{overall-table}, AdaPRL demonstrated a significant improvement over the traditional L2 loss function across the majority of regression datasets. However, when comparing AdaPRL employed directly as $ L_{reg}$ with the RankSim and Ordinal Entropy methods, AdaPRL performed slightly inferiorly on certain datasets such as KuaiRec, AgeDB and Jigsaw. Consequently, it was essential to explore whether integrating RankSim and Ordinal Entropy with AdaPRL could further boost the performance. Therefore, we designed new experiments on these three datasets by combining AdaPRL with both RankSim and Ordinal Entropy, aiming to assess the resultant performance.

Specifically, the hyperparameters of both RankSim and Ordinal Entropy method were optimized in each dataset via a coarse-grained grid search, and the weights were searched across \{0.05,0.1,0.2,0.4,0.8,1.6\}. Secondly, as we described in algorithm\ref{alg:adaprl_algorithm}, we computed and added an alpha-weighted $L_{CPRL}$ to the loss function of RankSim and Ordinal Entropy with the corresponding optimized hyperparameter. Then a coarse-grained grid search was performed on $\alpha$ of AdaPRL. The grid search and other settings were kept the same as the earlier experiment details.
\subsubsection{Results}
The results are presented in Table \ref{AdaPRL-rs-oe}. As shown in the table, the integration of AdaPRL with both RankSim and Ordinal Entropy exhibited performance improvement compared to those without AdaPRL in the KuaiRec, AgeDB as well as Jigsaw datasets. Specifically, for RankSim method, the improvement of combining AdaPRL in MSE and MAE metrics was 0.03\%, 1.04\% for KuaiRec, 0.77\%, 0.17\% and 0.09\% for AgeDB, and 0.23\% and 5.46\% for Jigsaw. The improvement for Ordinal Entropy was 6.46\% and 16.72\% for KuaiRec, 1.15\% and 0.66\% for AgeDB, and 2.22\% and 2.57\% for Jigsaw, respectively. The results demonstrate that our proposed AdaPRL is not a substitute for these existing techniques. Instead, it can be seamlessly integrated into various regression frameworks to achieve further performance improvements.

\begin{table*}[h]
\vskip 0.15in
\begin{center}
\renewcommand{\arraystretch}{1.2} 
\begin{small}
\begin{sc}\resizebox{0.925\textwidth}{!}{%
\begin{tabular}{l|ccc|ccc|ccc}
\toprule
Dataset & \multicolumn{3}{c|}{KuaiRec} & \multicolumn{3}{c|}{AgeDB} & \multicolumn{3}{c}{Jigsaw} \\ 
\cmidrule(lr){1-10}
Model & MSE$\downarrow$ & MAE$\downarrow$ & Kendall's \(\tau\)$\uparrow$ & MSE$\downarrow$ & MAE$\downarrow$ & Kendall's \(\tau\)$\uparrow$ & MSE$\downarrow$ & MAE$\downarrow$ & Kendall's \(\tau\)$\uparrow$ \\ 
\midrule
RankSim & 0.9338 & 0.2700 & 0.5761 & 68.938 & 6.368 & 0.6976 & 0.344 & 0.304 & 0.687 \\ 
\textbf{RankSim + AdaPRL} & \textbf{0.9335} & \textbf{0.2672} & \textbf{0.5808} & \textbf{68.405} & \textbf{6.357} & \textbf{0.6982} & \textbf{0.343} & \textbf{0.288} & \textbf{0.694} \\ 
\midrule
\emph{Rel.Imp} & 0.03\% & 1.04\% & 0.82\% &0.77\% & 0.17\% &0.09\% & 0.23\% & 5.46\% & 1.04\% \\
\midrule
Ordinal Entropy & 1.1825 & 0.4122 & 0.4699 & 68.873 & 6.380 & 0.6985 & 0.363 & 0.289 & 0.690 \\ 
\textbf{Ordinal Entropy + AdaPRL} & \textbf{1.1061} & \textbf{0.3433} & \textbf{0.4895} & \textbf{68.079} & \textbf{6.338} & \textbf{0.6993} & \textbf{0.355} & \textbf{0.282} & \textbf{0.691} \\ 
\midrule
\emph{Rel.Imp} & 6.46\% & 16.72\% & 4.17\% & 1.15\% & 0.66\% & 0.11\% & 2.22\% & 2.57\% & 0.20\% \\
\bottomrule
\end{tabular}%
}
\end{sc}
\end{small}
\caption{Comparison of performance of RankSim, Ordinal Entropy, and their AdaPRL-integrated methods on the KuaiRec, AgeDB, and Jigsaw datasets. \emph{Rel.Imp} denotes the relative improvement over the corresponding metrics. The experiments were repeated five times and the average results were reported.}
\label{AdaPRL-rs-oe}
\end{center}
\vskip -0.1in
\end{table*}

\subsection{Robustness to Label Noise with AdaPRL}
\subsubsection{Experiment Setup}
Label noise is a common problem in real-world datasets, where mislabeling can occur due to data ambiguity, human error, automated labeling processes, or intrinsic randomness of data, which can significantly degrade the performance of deep neural networks because of its misleading effects. As we mentioned before, AdaPRL could theoretically alleviate aleatoric noise of the data because of the uncertainty estimation module. To prove the theory, we evaluated our method against the standard baseline MSE loss using the KuaiRec dataset under varying levels of Gaussian noise for labels. The noise was introduced into the labels at different levels, where the label noise level $k$ signified that Gaussian noise with zero mean and a standard deviation of $0.2k \times \mathrm{std}(y)$ was added to the labels, and $\mathrm{std}(y)$ denoted the standard deviation of the original labels. This setup allowed us to assess the robustness of our method in the presence of different degrees of label noise. The experiment was repeated five times and the average mean squared error (MSE) of the test data was reported. 
\subsubsection{Results}
The result is presented in Figure \ref{sparsity-ett}. As we can see, AdaPRL consistently outperformed the baseline, exhibiting significantly less performance degradation as the proportion of noisy labels increases. Furthermore, as the level of label noise increases, our proposed method exhibits greater relative improvements in the MSE metric compared to the traditional MSE loss. This indicates that our method effectively mitigates the impact of label noise, resulting in more reliable model performance even when training data labels are imperfect.

\begin{figure}[ht]
\vskip 0.2in
\begin{center}
\begin{minipage}{1.0\columnwidth}
\centerline{\includegraphics[width=\columnwidth]{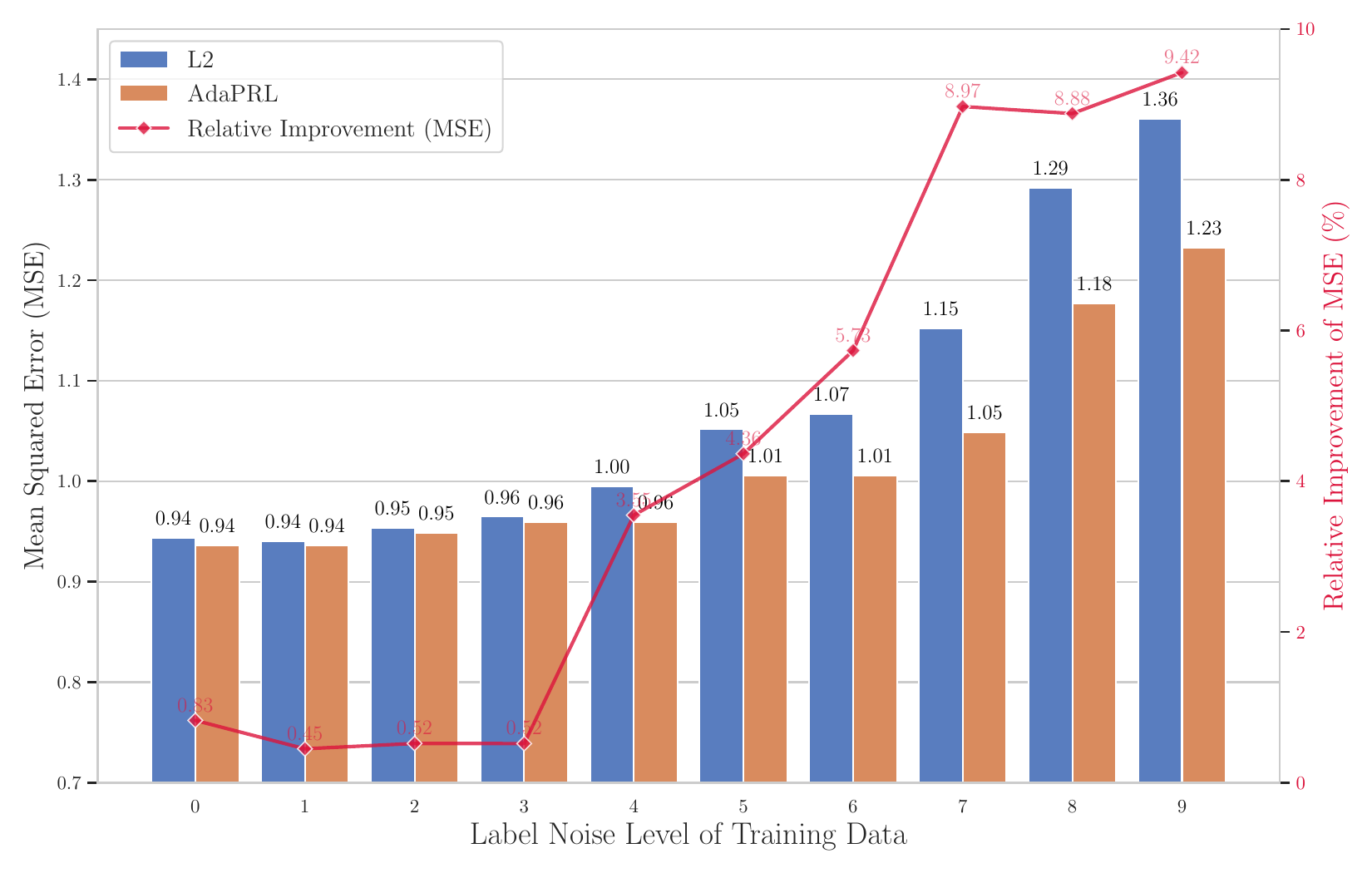}}
\caption{Performance comparison of AdaPRL and L2 loss on MSE metrics with different levels of label noise in training data on the KuaiRec dataset.}
\label{kr-label-noise}
\end{minipage}
\end{center}
\vskip -0.2in
\end{figure}

\subsection{Robustness to Unforeseen Data Corruption using AdaPRL}
\subsubsection{Experiment Setup}
Ensuring that deep neural networks maintain performance in the presence of unforeseen data corruptions in the inferring stage is essential for their deployment in real-world applications. Models often encounter various unexpected corruptions or inconsistencies, such as missing data, noise from data collection, and inconsistencies of online and offline features, which could significantly degrade online performance. To simulate the problems, we chose the Movielens-1M and KuaiRec datasets as representative, and the evaluation protocol is as follows. Firstly, we randomly selected 20\% of the columns to corrupt. Secondly, for each corruption level $k$, selected $10k\%$ of the validation and test samples. Then, randomly shuffled the values within the selected samples for each column. Finally, trained the model with the original training data and evaluated it on the corrupted test set. Note that in this experiment, our model was trained solely on uncorrupted data but was evaluated on corrupted test sets. Early stoppings were applied on the original validation data, and a coarse-grained grid search over alpha was conducted as mentioned above. The experiment was repeated five times, and the average of mean squared error (MSE) of the test data was reported.

\begin{figure}[ht]
\vskip 0.2in
\begin{center}
\begin{minipage}{1.0\columnwidth}
\centerline{\includegraphics[width=\columnwidth]{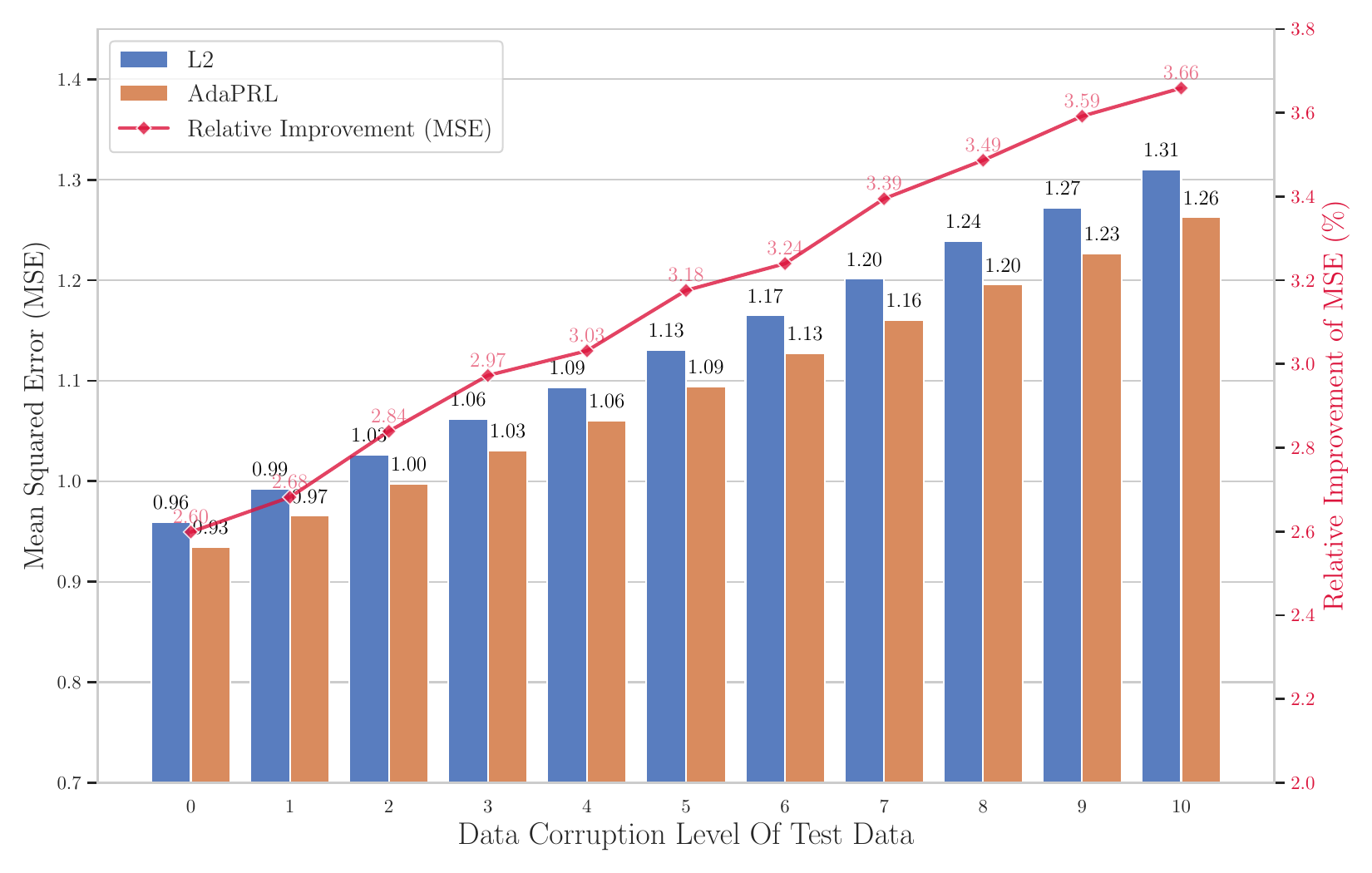}}
\caption{Performance comparison of AdaPRL and L2 loss on MSE metrics with different levels of unforeseen random corruption in test data on the KuaiRec dataset.}
\label{kr-test-corrupt}
\end{minipage}
\end{center}
\vskip -0.2in
\end{figure}

\subsubsection{Results}
As shown in Figure \ref{kr-test-corrupt}, our proposed method AdaPRL achieved evident improvement over the baseline and consistently achieved relative reductions in MSE compared to the L2 loss as the severity of data corruption increased. Furthermore, the relative improvement in MSE gradually increases as the level of data corruption increases, thereby demonstrating the improved resilience of our model to unforeseen test data corruptions.

\begin{figure}[ht]
\vskip 0.2in
\begin{center}
\begin{minipage}{1.0\columnwidth}
\centerline{\includegraphics[width=\columnwidth]{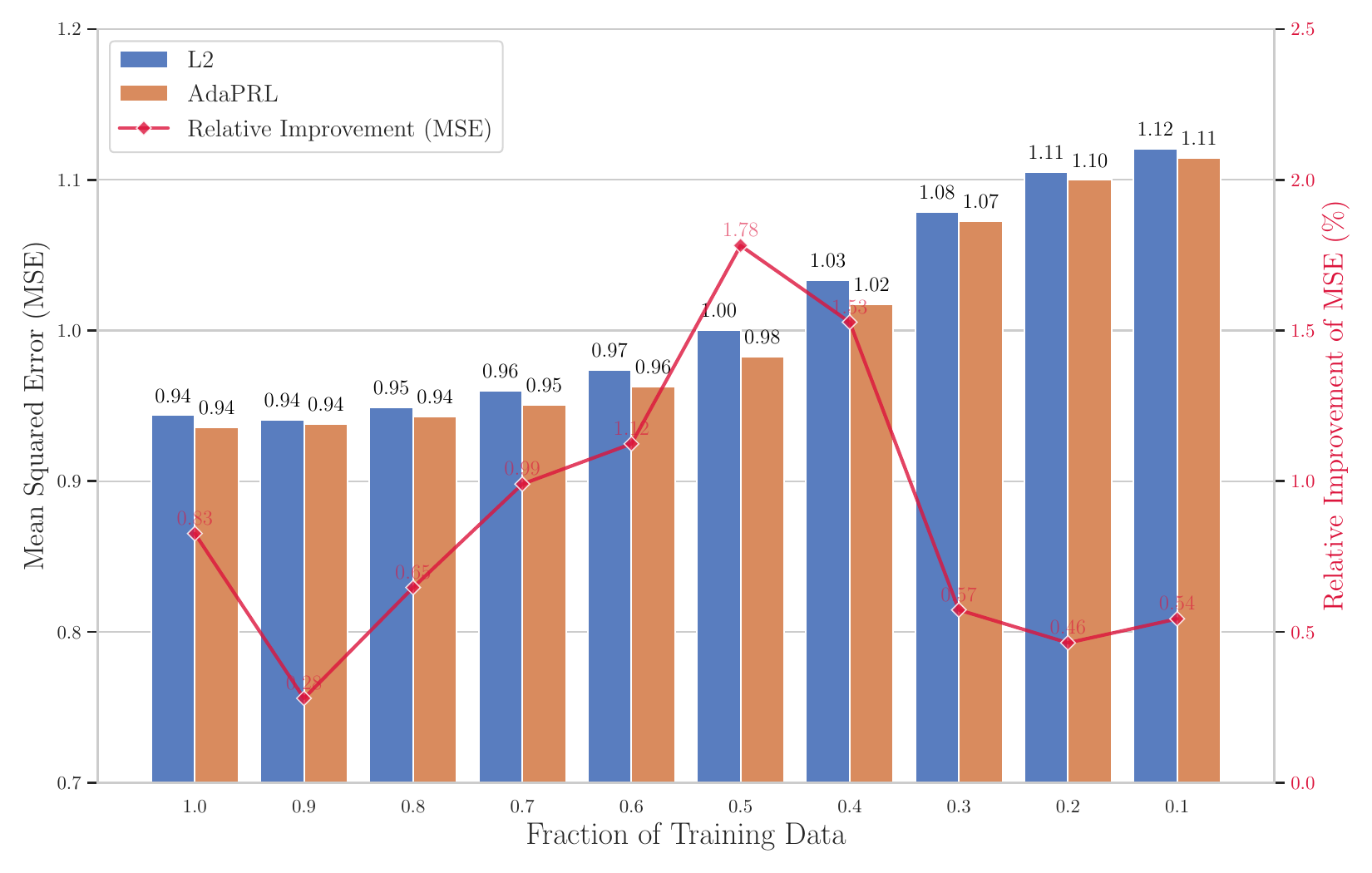}}
\caption{Performance comparison of AdaPRL and L2 loss on MSE metrics with different fractions of in training data on KuaiRec dataset.}
\label{kr-data-frac}
\end{minipage}
\end{center}
\vskip -0.2in
\end{figure}

\begin{figure*}[thbp]
    \centering
    \includegraphics[width=0.495\textwidth]{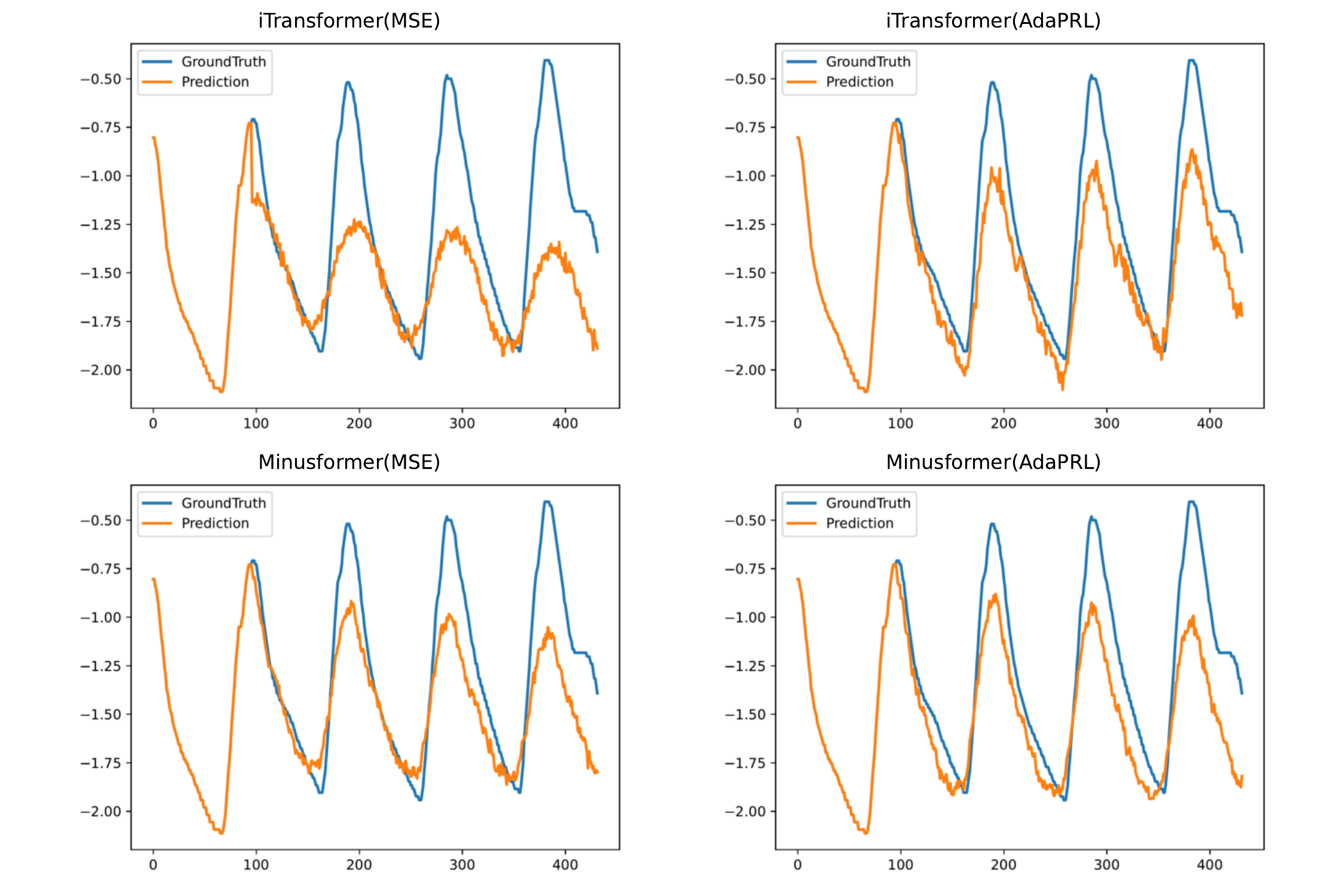}
    \includegraphics[width=0.495\textwidth]{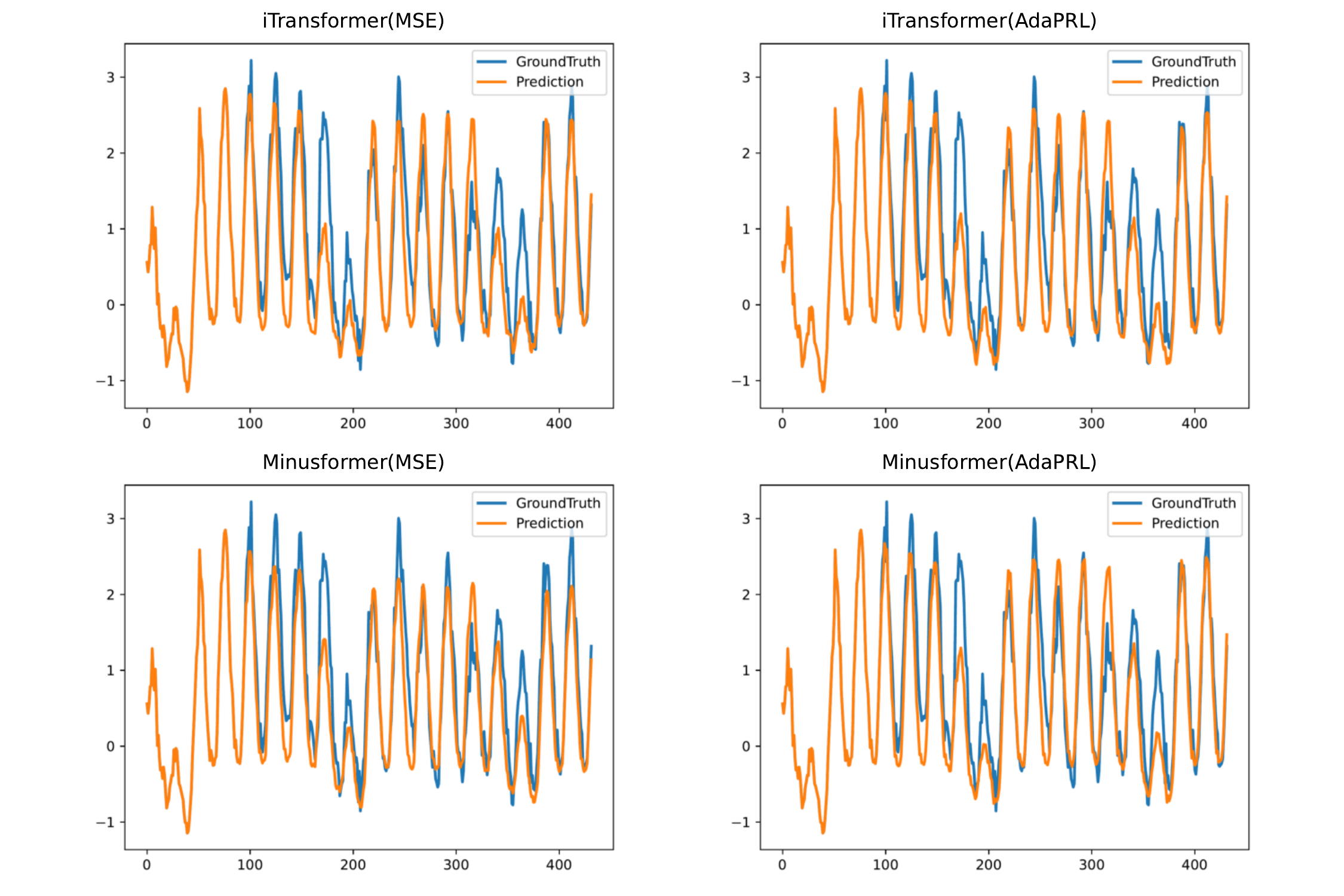}
     \\
    \makebox[0.495\textwidth]{\small (a) ETTm2 Dataset Results.}
    \makebox[0.495\textwidth]{\small (b) ECL Dataset Results.}
    \caption{Comparison of input-96-predict-336 results on ETTm2 and ECL datasets.}
    \label{vis-ettm2-ecl}
\end{figure*}

\subsection{Resilience to Reduced Training Data with AdaPRL}
\subsubsection{Experiment Setup}
Although the success of deep learning models is heavily dependent on the scale of training datasets, acquiring large quantities of labeled data is often impractical due to resource constraints. In many real-world situations, models must be effective even with limited training data. In this section, we conducted experiments by creating random subsets of the KuaiRec dataset with varying fractions to simulate scenarios with limited data. Our method was compared against a standard L2 loss baseline under the same conditions. The experiment was repeated five times and the average mean squared error (MSE) of test data was reported.

\subsubsection{Results}
As shown in Figure \ref{kr-data-frac}, our model demonstrates superior performance with significantly less degradation as the number of training samples decreases. This indicates that AdaPRL is more effective in settings where training data is limited, making it a viable solution for applications with constrained data availability. Another conclusion that can be drawn is that AdaPRL outperformed L2 loss in all cases and exhibited the highest relative improvement when the number of training data was at a moderate level.

\subsection{Visualization of Time Series Forecasting Prediction Results }\label{vis-explain-section}

To facilitate an instinct comparison among different backbone forecasting models with or without AdaPRL, we present prediction visualizations for two representative datasets in Figure \ref{vis-ettm2-ecl}. The figure illustrated the prediction results produced by the following models: iTransformer \cite{liu_itransformer_2024} and Minusformer \cite{liang_minusformer_2024} with a prediction length of 336. Upon visual inspection of the prediction results across these datasets, both iTransformer and Minusformer trained with AdaPRL demonstrated better accuracy in capturing future series variations, exhibiting superior performance in long-term forecasting compared to using MSE loss, which was consistent with the previous experimental result.

 \begin{center}
 \begin{figure}[ht]
 \vskip 0.1in
 \begin{center}
 \begin{minipage}{0.975\columnwidth}
 \centerline{\includegraphics[width=\columnwidth]{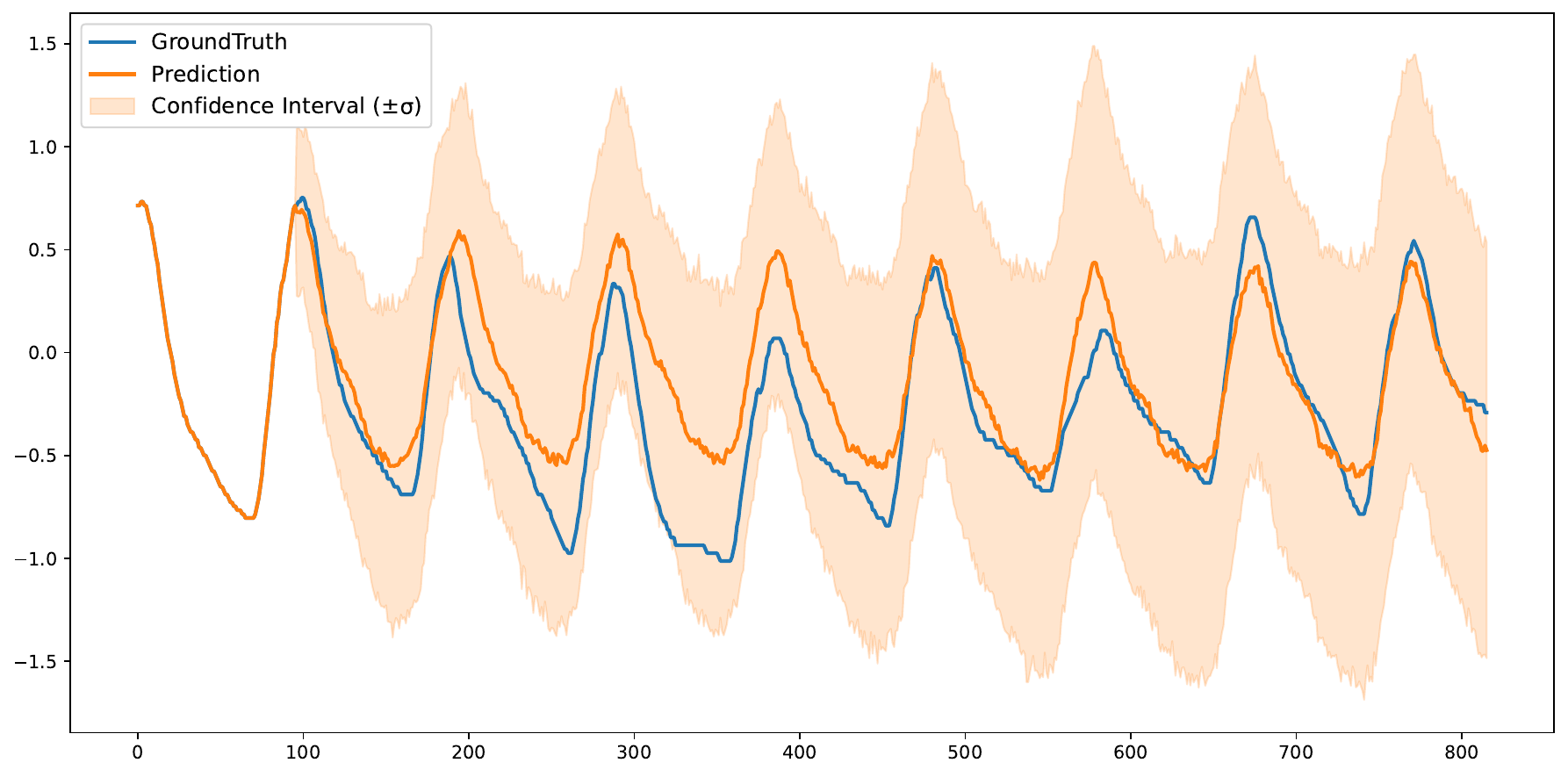}}
 \caption{Visualization of input-96-predict-720 results on the ETTm2 dataset using AdaPRL method with Minusformer as backbone network. This interval represented the confidence interval corresponding to standard error above and below the prediction expectation.}
 \label{vis-explain-ettm2}
 \end{minipage}
 \end{center}
 \vskip -0.1in
 \end{figure}
 \end{center}

\subsection{Visualization of Time Series Forecasting Uncertainty Estimation }
By integrating a deep probabilistic network with an arbitrary backbone network, our proposed AdaPRL is capable of generating not only point estimates for each timestamp in long-term time series forecasting but also predicting the associated uncertainties. Figure \ref{vis-explain-ettm2} is an example from the ETTm2 dataset with a prediction length of 720. The figure demonstrates that the orange interval is defined by adding and subtracting the standard deviation $\sigma$ values predicted by the deep probabilistic network. This interval represents the confidence interval corresponding to one standard error above and below the prediction expectation. Furthermore, as is illustrated in this figure, the uncertainty levels increase as the prediction horizon extends. This trend is consistent with expectations, as the model faces greater uncertainty when forecasting events further into the future.

Without loss of generality, AdaPRL enhances the interpretability of prediction results by providing uncertainty estimations for each prediction value across various types of regression tasks. This capability is particularly valuable as it allows practitioners to gauge the reliability of predictions, thereby facilitating more informed decision-making processes. By incorporating uncertainty quantification, AdaPRL not only predicts the target values but also offers insights into the confidence associated with these predictions, which is crucial in applications where the precision and trustworthiness of predictions are important.

 \begin{center}
 \begin{figure}[ht]
 \vskip 0.1in
 \begin{center}
 \begin{minipage}{1.0\columnwidth}
 \centerline{\includegraphics[width=\columnwidth]{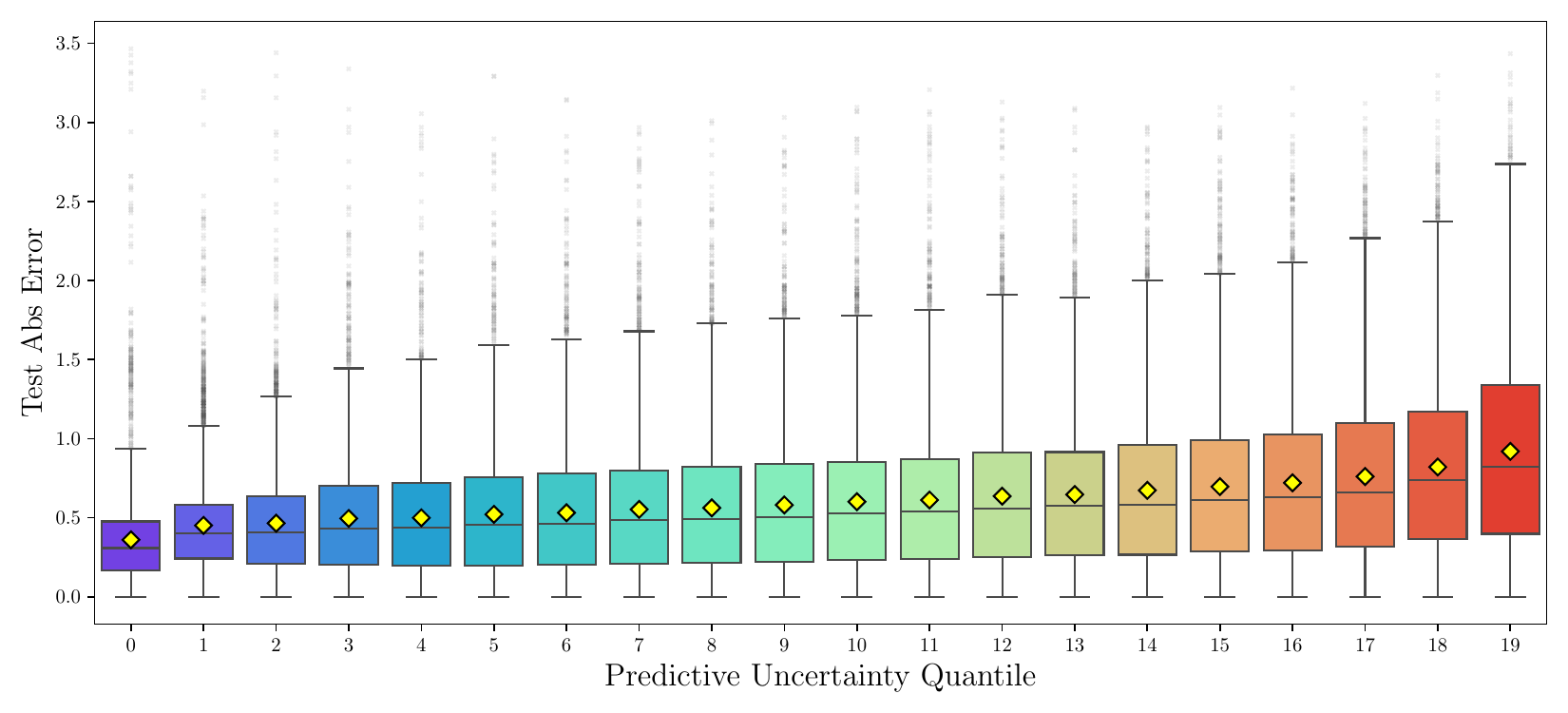}}
 \caption{Box plot of predictive uncertainty and test absolute error using AdaPRL method in the test set of Movielens-1M dataset.}
 \label{boxplot-ml1m}
 \end{minipage}
 \end{center}
 \vskip -0.1in
 \end{figure}
 \end{center}

 \begin{center}
 \begin{figure}[ht]
 \vskip 0.1in
 \begin{center}
 \begin{minipage}{1.0\columnwidth}
 \centerline{\includegraphics[width=\columnwidth]{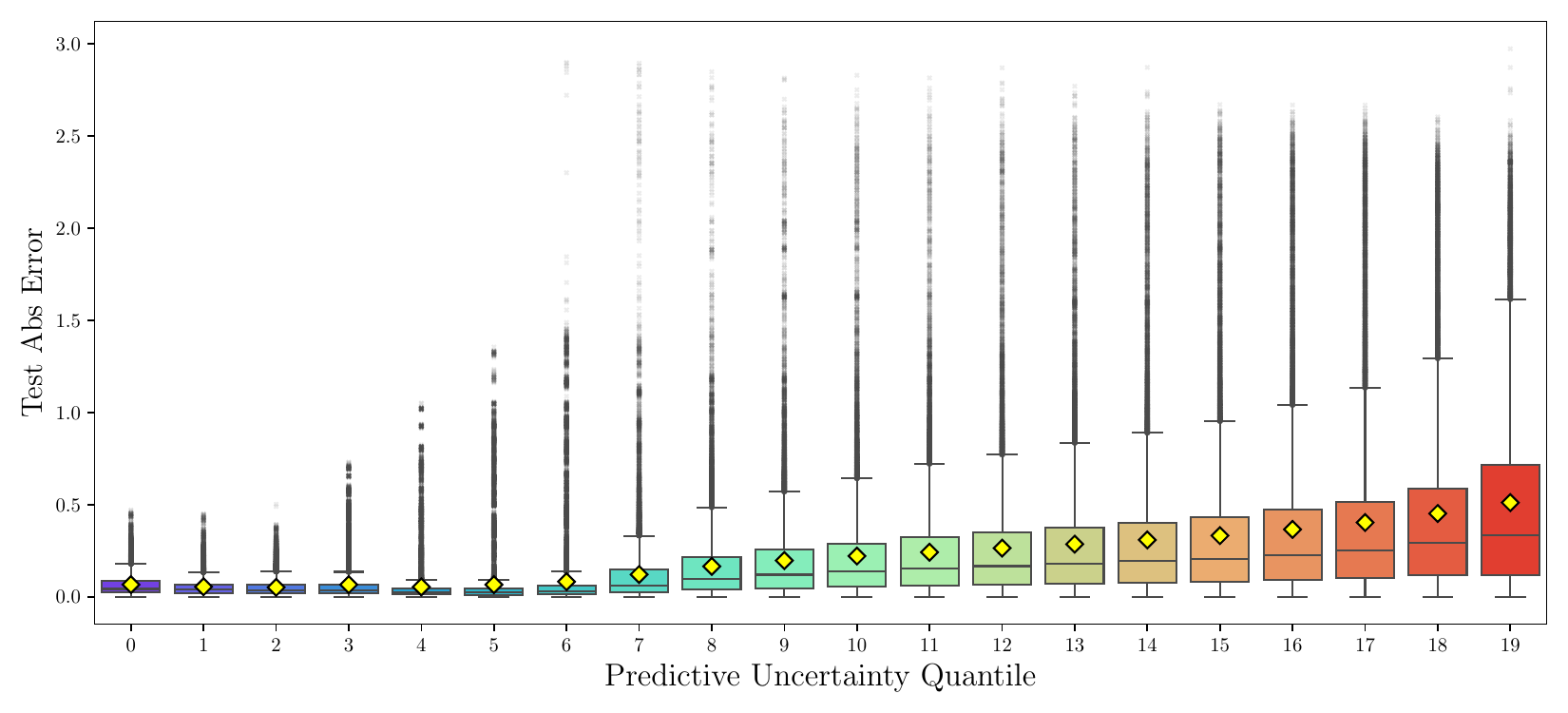}}
 \caption{Box plot of predictive uncertainty and test absolute error using AdaPRL method in the test set of KuaiRec dataset.}
 \label{boxplot-kr}
 \end{minipage}
 \end{center}
 \vskip -0.1in
 \end{figure}
 \end{center}

\subsection{Analysis of Auxiliary Deep Probabilistic Network}
\subsubsection{Experiment Setup}
To better understand the performance of the auxiliary deep probabilistic network as well as the relationship between the out-of-fold uncertainty estimation and prediction errors, we conducted the following experiments on the Movielens-1M and KuaiRec datasets. First, we trained the model using AdaPRL on the training set following former experimental setups. Then, using the auxiliary deep probabilistic network of our model, we made predictions on the test set, obtaining uncertainty estimates $\sigma$ for each sample in the test set, and we obtained the point estimation from the main network. Next, we calculated the point estimate test error for the means and compared the correlation between the standard deviations and the actual test errors. 
\subsubsection{Results}
The results are shown in Figures \ref{boxplot-ml1m} and \ref{boxplot-kr}. The evaluation result of the test set was grouped according to the uncertainty estimation predicted by the deep probabilistic network in AdaPRL, and the absolute error distribution of each group was on the y-axis. The diamond signal indicated the average test error of each group. Note that the outliers of the KuaiRec dataset were eliminated for better illustration. As shown in the figures, when the model's predicted standard deviation was larger, indicating higher uncertainty in the model's prediction for these samples, we indeed observed that the actual test error for that sample tended to be larger, showing an obvious positive correlation between the two. The phenomenon is very similar in both datasets, which clearly demonstrates the effectiveness of the deep probabilistic network of AdaPRL in estimating uncertainties.

\begin{table*}[h]
\vskip 0.15in
\begin{center}
\renewcommand{\arraystretch}{1.2} 
\begin{small}
\begin{sc}\resizebox{0.9\textwidth}{!}{%
\begin{tabular}{c|c|cc|c|cc|c}
\toprule
\multicolumn{2}{c}{Dataset}  & \multicolumn{3}{|c|}{Movielens-1M} & \multicolumn{3}{c}{KuaiRec} \\ 
\midrule
Training Platform & Batch Size & time(MSE)  & time(AdaPRL)  & \emph{Rel.Inc}  & time(MSE)  & time(AdaPRL)  & \emph{Rel.Inc}   \\ 
\midrule
\multirow{5}{*}[-1.5ex]{
    \begin{tabular}{c}
      \textbf{CPU} \\
      \tiny{Intel Xeon Platinum 8168}
    \end{tabular}
}
& 256 & 27.27 & 49.24 & 80.6\% & 496.15  & 978.54 & 97.2\% \\ 
& 512 & 16.53  & 30.96 & 87.3\% & 325.36  & 650.23 & 99.9\% \\ 
& 1024 & 11.06 & 20.60 & 86.3\% & 258.51  & 471.60 & 82.4\% \\ 
& 2048 & 8.63 & 31.58 & 265.8\% & 206.90  & 449.38 & 117.2\% \\ 
& 4096 & 8.77 & 79.25 & 804.0\% & 195.70  & 701.65 & 258.5\% \\ 
& 8192 & 8.84 & 150.24 & 1600.3\% & 197.84  & 1111.05 & 461.1\% \\ 
\midrule
\multirow{5}{*}[-1.5ex]
{
    \begin{tabular}{c}
      \textbf{GPU} \\
      \tiny {NVIDIA RTX A6000}
    \end{tabular}
}
& 256 & 13.64 & 27.99 & 105.2\% & 257.66  & 527.47 & 104.7\% \\ 
& 512 & 9.04 & 15.73 & 74.0\% & 174.36  & 260.12 & 49.2\% \\ 
& 1024 & 6.12 & 10.76 & 75.9\% & 125.23  & 170.51 & 36.2\% \\ 
& 2048 & 6.34 & 8.21 & 29.5\% & 106.69  & 143.14 & 34.2\% \\ 
& 4096 & 7.41 & 8.88 & 19.9\% & 113.99  & 142.30 & 24.8\% \\ 
& 8192 & 7.76 & 9.01 & 16.2\% & 106.23  & 123.08 & 15.9\% \\ 

\bottomrule
\end{tabular}%
}
\end{sc}
\end{small}
\caption{Training time cost each epoch on Movielens-1M and KuaiRec datasets using MLP as backbone model. The benchmark experiments were repeated 5 times and the minimum of corresponding training time cost were reported. \emph{Rel.Inc} indicates the relative increase of training time cost of AdaPRL over the baseline MSE loss.}
\label{time-cost-exp}
\end{center}
\vskip -0.1in
\end{table*}

\subsection{Theoretical And Experimental Analysis of Extra Computation Costs with AdaPRL}

\subsubsection{Theoretical Analysis}
Earlier experiments have shown that AdaPRL consistently improves the precision of regression tasks as well as the robustness of regression models. Despite all those advantages, it is necessary to conduct an analysis of its computational cost both theoretically and experimentally in order to ensure its efficiency and scalability for large-scale applications. 

During the inference stage, if the point estimation is the only concern, as is typically the case in many real-world applications, then the auxiliary network can be ignored and only the original network needs to be deployed; therefore, the inference cost remains exactly the same as the original model, and there will be \emph{no} extra computational cost during inference at all.

During the training stage, however, extra computation is inevitable for two reasons. Firstly, in any single-output pairwise regression learning (PRL) framework, the difference of the pairwise predictions, the difference of the pairwise labels, as well as their difference are necessary to be calculated, resulting in extra computation cost. Assume the mini-batch size is $B$ then the extra computing cost should be around $O(B^2)$ for each step. Secondly, since an auxiliary deep probabilistic network is introduced, extra computing is also needed during the training phase in both the network forward and backward processes. If the same backbone is used as the auxiliary network as we did in our experiments, then the training computing task is approximately twice as much as using a single network if the slight difference of the output layer is ignored. 

In conclusion, when the batch size in the training phase is small, the additional computational overhead of the auxiliary network is the dominant factor, leading to an approximate doubling of the overall computational cost compared to the original model theoretically. As the batch size increases, however, the extra cost associated with pairwise computations becomes the primary contributor, causing the additional theoretical computational cost to grow proportionally to the square of the batch size.

\subsubsection{Experiment Setup}
Besides the theoretical analysis, we also conducted multiple comparative benchmark experiments on Movielens-1M and KuaiRec datasets to facilitate a better understanding of the additional time cost of training AdaPRL compared to the baseline loss in actual applications. We trained the models using either CPU or GPU platforms with different batch sizes to assess the overall speed performance differences across hardware platforms. The experiment was run with PyTorch 2.4.1 and CUDA 12.4 environment. For other detailed information of our hardware platform and their corresponding results, please refer to Table \ref{time-cost-exp}.
\subsubsection{Experimental Results and Analysis}

The outcomes of these experiments are depicted in Table \ref{time-cost-exp}. As shown in the table, in the context of CPU-based model training, our experimental findings across both datasets aligned with the theoretical analysis. Specifically, for smaller batch sizes ($\leq$1024), the training time for AdaPRL was approximately 1.8 to 2.0 times that of the MSE loss function. Conversely, for larger batch sizes ($\geq$2048), the relative increase in training time for AdaPRL compared to MSE exhibited near-quadratic growth, as the additional computational cost tended to increase roughly with the square of the batch size. These benchmark results confirm the anticipated performance attributes of AdaPRL.

In contrast, GPU-accelerated training yielded different results due to hardware optimizations. For smaller batch sizes (e.g., 256), AdaPRL's training time was also approximately double that of MSE. However, as batch sizes increased, the relative increment in training time per epoch decreased dramatically from around 105\% to around 16\%. Consequently, when training regression models for tabular data at moderate batch sizes with GPU, the computational time increase when using AdaPRL instead of MSE is expected to be around 30\% and 70\%. Given that no additional computational costs are introduced during inference, the computational overhead introduced by AdaPRL is generally considered acceptable in practical industrial scenarios.

\section{Conclusion and Further Study}
As we have analyzed in previous sections, our proposed AdaPRL method demonstrates notable potential in several key areas. Firstly, AdaPRL has the ability to enhance prediction accuracy and ranking ability in various domains according to our extensive experiments, which are crucial for improving the performance of modern deep learning regression models. Secondly, AdaPRL has contributed to increasing generalization capability, enabling the model to perform better on unforeseen data corruption and limited data scenarios. In addition, it has exhibited improved robustness to noisy data, which is considered particularly important in real-world applications. Moreover, our proposed AdaPRL enhances interpretability because it allows practitioners to gain deeper insights into the model's decision-making process and fosters trust in its predictions. Lastly, our experiments show that AdaPRL can seamlessly incorporate into different regression methods, such as RankSim and Ordinal Entropy, yielding even better accuracy. In conclusion, these combined advantages position our approach as a promising improved solution in the field of regression tasks.

Therefore, we believe that this method extends to a broader range of deep learning problems, particularly classification tasks. Uncertainty-based pairwise learning demonstrates exceptional performance improvements and generalization enhancements for regression tasks, leading us to believe that similar benefits apply to classification problems. This represents a research direction with great potential; however, since this paper primarily focuses on regression problems, we do not delve further into this topic here. Instead, we intend to explore this area in our future studies.

\end{document}